\documentclass[10pt,twocolumn,letterpaper]{article}

\usepackage{cvpr}              % To produce the CAMERA-READY version
\makeatletter
\@namedef{ver@everyshi.sty}{}
\makeatother
\usepackage{tikz}
\usepackage{graphicx}
\usepackage{amsmath}
\usepackage{amssymb}
\usepackage{booktabs}
\usetikzlibrary{calc}
\usepackage{tablefootnote}
\usepackage{csquotes}
\usepackage{enumerate}
\usepackage{paralist}

\usepackage{caption}
\usepackage{subcaption}

\usepackage[pagebackref,breaklinks,colorlinks]{hyperref}

\usepackage[capitalize]{cleveref}
\crefname{section}{Sec.}{Secs.}
\Crefname{section}{Section}{Sections}
\Crefname{table}{Table}{Tables}
\crefname{table}{Tab.}{Tabs.}

\usepackage{pifont}
\newcommand{\xmark}{\ding{55}}
\newcommand{\cmark}{\ding{51}}
\usepackage{adjustbox}

\usepackage{multirow}
\usepackage{array}
\usepackage{colortbl}
\newcolumntype{C}[1]{>{\centering\arraybackslash}m{#1}}
\newcolumntype{G}[2]{>{\columncolor[gray]{0.90}[#1][#2]}c}

\def\cvprPaperID{2466} % *** Enter the CVPR Paper ID here
\def\confName{CVPR}
\def\confYear{2023}

\newcommand*\splitimgslant{0.05}
\newcommand*\splitimggap{1pt}
\newcommand*{\splitimgshifts}[5]{%
\begin{tikzpicture}%
\node (rect) at (0,0) [outer sep=0,anchor=south west,draw=black!50,line width=.25pt,fill=none,opacity=0,minimum width=#1,minimum height=#1/1920*1080] {};
\begin{scope}
\path[clip] ([xshift=\splitimgslant*#1-0.5*\splitimggap]rect.north) -- ([xshift=-\splitimgslant*#1-0.5*\splitimggap]rect.south) -- (rect.south west) -- (rect.north west) -- cycle;
\node[inner sep=0,anchor=south west,outer sep=0] (A) at (#4*#1,0) {\includegraphics[width=#1]{#2}};
\end{scope}
\begin{scope}
\path[clip] ([xshift=\splitimgslant*#1+0.5*\splitimggap]rect.north) -- ([xshift=-\splitimgslant*#1+0.5*\splitimggap]rect.south) -- (rect.south east) -- (rect.north east) -- cycle;
\node[inner sep=0,anchor=south west,outer sep=0] (B) at (#5*#1,0) {\includegraphics[width=#1]{#3}};
\end{scope}
\node (rect) at (0,0) [outer sep=0,anchor=south west,draw=black!50,line width=.25pt,fill=none,opacity=1,minimum width=#1,minimum height=#1/1920*1080] {};
\end{tikzpicture}
}

\newcommand*{\splitimg}[3]{\splitimgshifts{#1}{#2}{#3}{0}{0}}

\newcommand*{\splitimgfourshifts}[9]{%
\begin{tikzpicture}%
\node (rect) at (0,0) [outer sep=0,anchor=south west,draw=black!50,line width=.25pt,fill=none,opacity=0,minimum width=#1,minimum height=#1/1920*1080] {};
\begin{scope}
\path[clip] ([xshift=\splitimgslant*#1-0.5*\splitimggap]$(rect.north west)!0.5!(rect.north)$) -- ([xshift=-\splitimgslant*#1-0.5*\splitimggap]$(rect.south west)!0.5!(rect.south)$) -- (rect.south west) -- (rect.north west) -- cycle;
\node[inner sep=0,anchor=south west,outer sep=0] (A) at (#6*#1,0) {\includegraphics[width=#1]{#2}};
\end{scope}
\begin{scope}
\path[clip] ([xshift=\splitimgslant*#1+0.5*\splitimggap]$(rect.north west)!0.5!(rect.north)$) -- ([xshift=-\splitimgslant*#1+0.5*\splitimggap]$(rect.south west)!0.5!(rect.south)$) -- ([xshift=-\splitimgslant*#1-0.5*\splitimggap]rect.south) -- ([xshift=\splitimgslant*#1-0.5*\splitimggap]rect.north) -- cycle;
\node[inner sep=0,anchor=south west,outer sep=0] (A) at (#7*#1,0) {\includegraphics[width=#1]{#3}};
\end{scope}
\begin{scope}
\path[clip] ([xshift=\splitimgslant*#1+0.5*\splitimggap]rect.north) -- ([xshift=-\splitimgslant*#1+0.5*\splitimggap]rect.south) -- ([xshift=-\splitimgslant*#1-0.5*\splitimggap]$(rect.south)!0.5!(rect.south east)$) -- ([xshift=\splitimgslant*#1-0.5*\splitimggap]$(rect.north)!0.5!(rect.north east)$) -- cycle;
\node[inner sep=0,anchor=south west,outer sep=0] (B) at (#8*#1,0) {\includegraphics[width=#1]{#4}};
\end{scope}
\begin{scope}
\path[clip] ([xshift=\splitimgslant*#1+0.5*\splitimggap]$(rect.north)!0.5!(rect.north east)$) -- ([xshift=-\splitimgslant*#1+0.5*\splitimggap]$(rect.south)!0.5!(rect.south east)$) -- (rect.south east) -- (rect.north east) -- cycle;
\node[inner sep=0,anchor=south west,outer sep=0] (B) at (#9*#1,0) {\includegraphics[width=#1]{#5}};
\end{scope}
\node (rect) at (0,0) [outer sep=0,anchor=south west,draw=black!50,line width=.25pt,fill=none,opacity=1,minimum width=#1,minimum height=#1/1920*1080] {};
\end{tikzpicture}
}

\newcommand*{\splitimgfour}[5]{\splitimgfourshifts{#1}{#2}{#3}{#4}{#5}{0}{0}{0}{0}}

\newcommand{\paragraphnew}[1]{\medskip\noindent\textbf{#1}}

\newcommand{\mapdetails}[7]{%
\begin{tikzpicture}%
\node[inner sep=0pt] (main) {\includegraphics[width=#1]{#2}};%
\node[inner sep=0pt, above left] (subord) at (main.south east) {\fcolorbox{red!70!white}{white}{\includegraphics[width=#3]{#4}}};%
\draw[draw=red!70!white,line width=.75pt] (#5,#6) rectangle (#5+.42,#6+.42) ; %0.42 boxwidth is ~190px rectangle%
\node[rectangle, inner sep=1.5pt, above right, fill=black, fill opacity=0.6, text opacity=1.] at (main.south west) {\footnotesize \textcolor{white}{#7}};
\end{tikzpicture}%
}

\newcommand{\darkrightbox}[3]{%
\begin{tikzpicture}
\node[inner sep=0pt] (p00) {\includegraphics[height=#1]{#2}};
\node[rectangle, inner sep=1.5pt, above left, fill=black, fill opacity=0.6, text opacity=1.] at (p00.south east) {\footnotesize \textcolor{white}{#3}}; \end{tikzpicture}%
}

\newcommand{\darklefttbox}[3]{%
\begin{tikzpicture}
\node[inner sep=0pt] (p00) {\includegraphics[width=#1]{#2}};
\node[rectangle, inner sep=1.5pt, above right, fill=black, fill opacity=0.6, text opacity=1.] at (p00.south west) {\footnotesize \textcolor{white}{#3}}; \end{tikzpicture}%
}

\newcommand{\darktoprightboxbox}[8]{%
\begin{tikzpicture}
\node[inner sep=0pt, anchor=south west] (p00) {\includegraphics[height=#1]{#2}};
\node[rectangle, inner sep=1.5pt, below right, fill=black, fill opacity=0.6, text opacity=1.] at (p00.north west) {\footnotesize \textcolor{white}{#3}};
\node[rectangle, inner sep=1.5pt, above left, fill=black, fill opacity=0.6, text opacity=1.] at (p00.south east) {\footnotesize \textcolor{white}{#4}};
\draw[draw=#5,line width=.5pt] (#6,#7) rectangle (#6+#8,#7+#8) ; %0.42 boxwidth is ~190px rectangle%
\end{tikzpicture}%
}

\newcommand{\lightrightbox}[3]{%
\begin{tikzpicture}
\node[inner sep=0pt] (p00) {\includegraphics[height=#1]{#2}};
\node[rectangle, inner sep=1.5pt, above left, fill=white, fill opacity=0.6, text opacity=1.] at (p00.south east) {\footnotesize \textcolor{black}{#3}}; \end{tikzpicture}%
}

\newcommand{\lighttoprightboxbox}[8]{%
\begin{tikzpicture}
\node[inner sep=0pt, anchor=south west] (p00) {\includegraphics[height=#1]{#2}};
\node[rectangle, inner sep=1.5pt, below right, fill=white, fill opacity=0.6, text opacity=1.] at (p00.north west) {\footnotesize \textcolor{black}{#3}};
\node[rectangle, inner sep=1.5pt, above left, fill=white, fill opacity=0.6, text opacity=1.] at (p00.south east) {\footnotesize \textcolor{black}{#4}};
\draw[draw=#5,line width=.5pt] (#6,#7) rectangle (#6+#8,#7+#8) ; %0.42 boxwidth is ~190px rectangle%
\end{tikzpicture}%
}

\begin{document}

%%%%%%%%% TITLE - PLEASE UPDATE
\title{Spring: A High-Resolution High-Detail Dataset and Benchmark\\for Scene Flow, Optical Flow and Stereo}

\author{Lukas Mehl \hspace{8mm} Jenny Schmalfuss \hspace{8mm} Azin Jahedi \hspace{8mm} Yaroslava Nalivayko \hspace{8mm} Andr\'es Bruhn\\
Institute for Visualization and Interactive Systems, University of Stuttgart\\
{\small{\tt firstname.lastname@vis.uni-stuttgart.de}}
}
\maketitle

%%%%%%%%% ABSTRACT
\begin{abstract}
While recent methods for motion and stereo estimation recover an unprecedented amount of details, such highly detailed structures are neither adequately reflected in the data of existing benchmarks nor their evaluation methodology.
Hence, we introduce Spring -- a large, high-resolution, high-detail, computer-generated benchmark for scene flow, optical flow, and stereo.
Based on rendered scenes from the open-source Blender movie \enquote{Spring}, it provides photo-realistic HD datasets with state-of-the-art visual effects and ground truth training data.
Furthermore, we provide a website to upload, analyze and compare results.
Using a novel evaluation methodology based on a super-resolved UHD ground truth, our Spring benchmark can assess the quality of fine structures and provides further detailed performance statistics on different image regions.
Regarding the number of ground truth frames, Spring is 60$\times$ larger than the only scene flow benchmark, KITTI 2015, and 15$\times$ larger than the well-established MPI Sintel optical flow benchmark.
Initial results for recent methods on our benchmark show that estimating fine details is indeed challenging, as their accuracy leaves significant room for improvement.
The Spring benchmark and the corresponding datasets are available at \href{http://spring-benchmark.org}{http://spring-benchmark.org}.
\end{abstract}

%%%%%%%%% BODY TEXT

\begin{figure}
\centering
\setlength\tabcolsep{1pt}
\def\arraystretch{0.0}
\begin{tabular}{@{}cc@{}}
\darktoprightboxbox{0.355\linewidth}{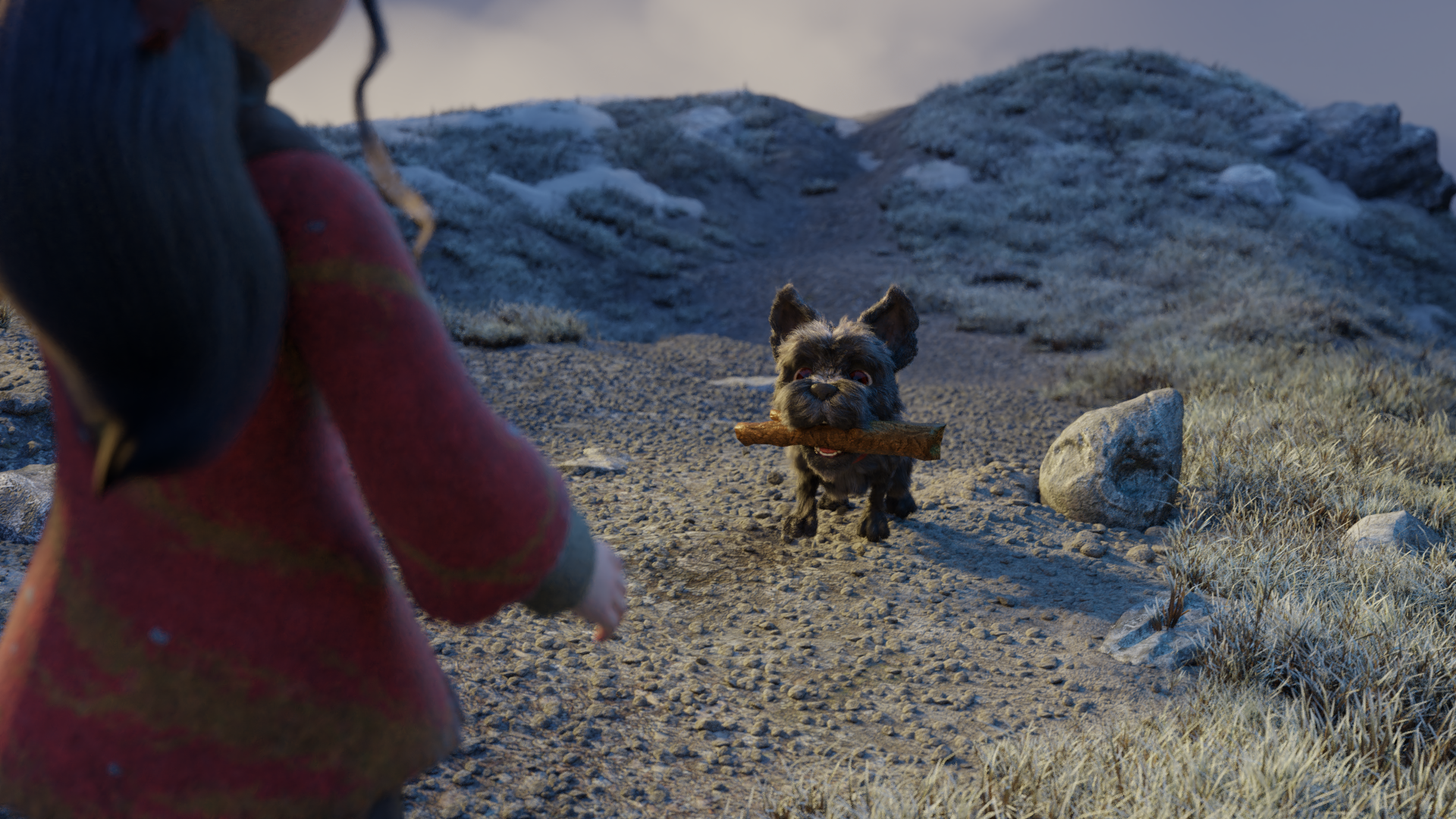}{Rendered Image}{1920$\times$1080}{black}{2.62}{1.1}{0.85} &
\darkrightbox{0.355\linewidth}{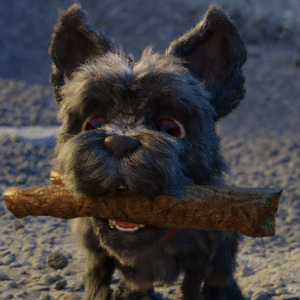}{Zoom 3.6$\times$}\\[2pt]
\lighttoprightboxbox{0.355\linewidth}{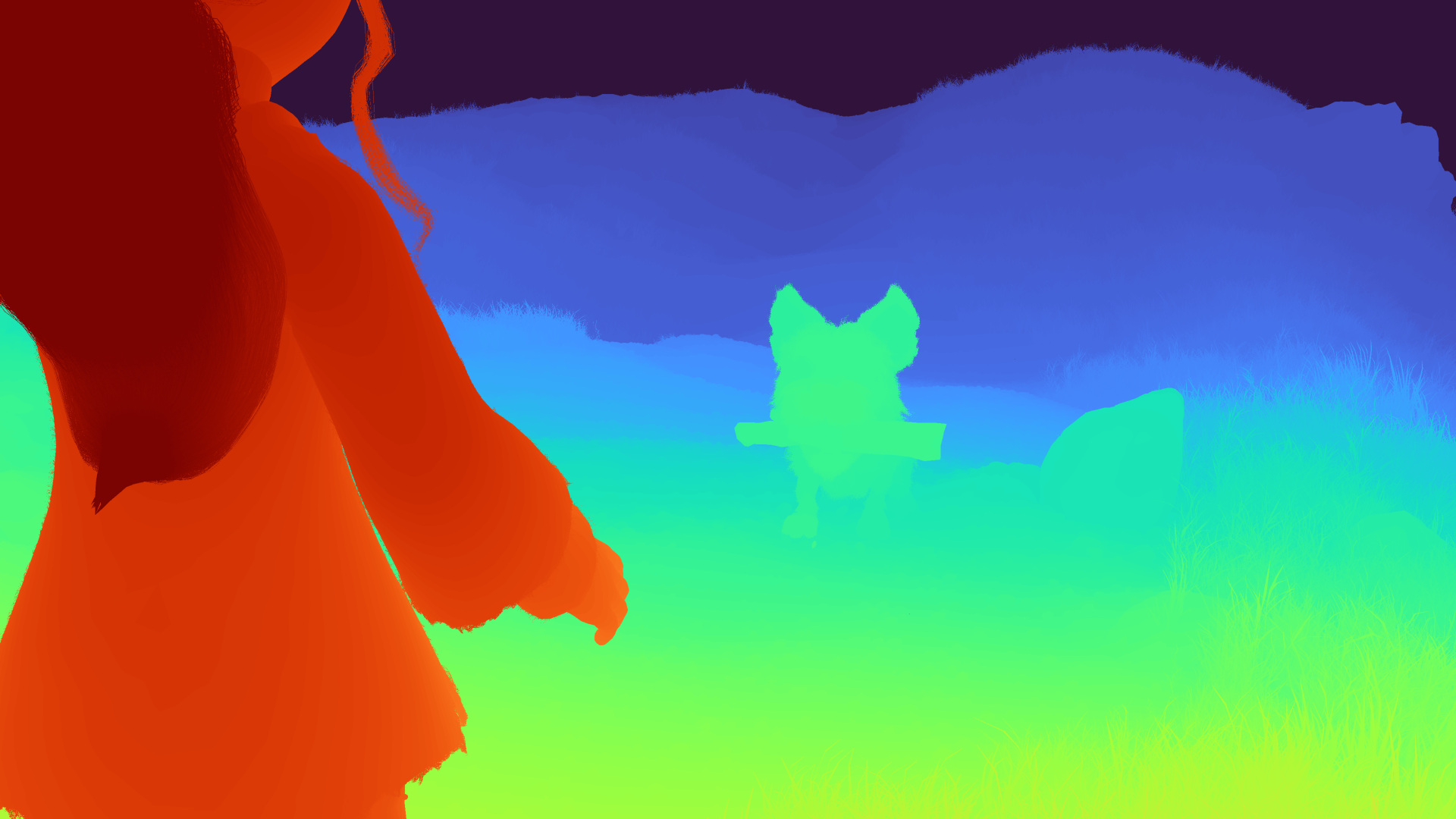}{Disparity}{3840$\times$2160}{black}{1.1}{2.15}{0.43} &
\lightrightbox{0.355\linewidth}{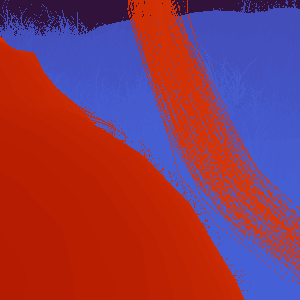}{Zoom 7.2$\times$}\\[2pt]
\lighttoprightboxbox{0.355\linewidth}{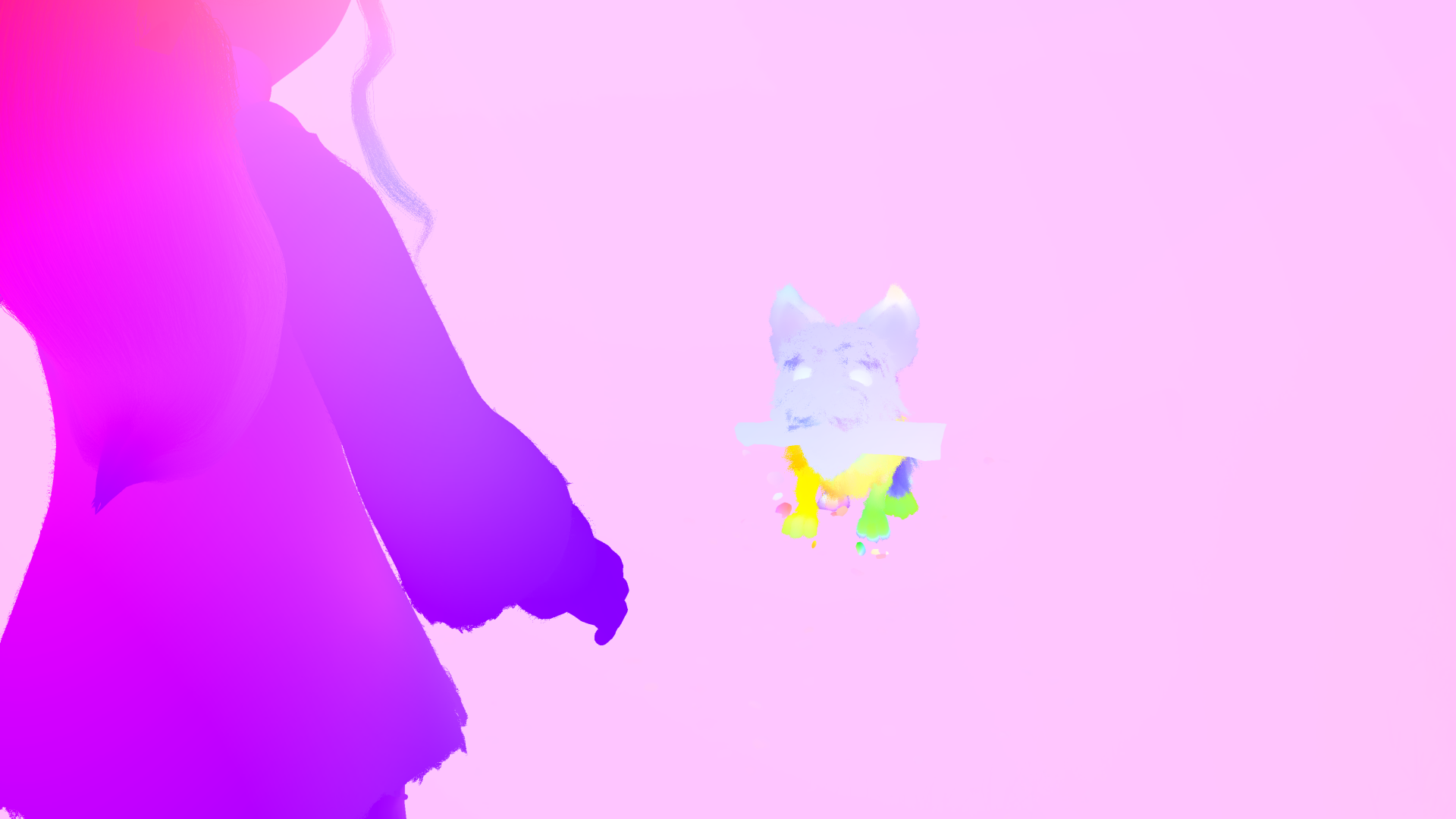}{Optical Flow}{3840$\times$2160}{black}{2.77}{.89}{0.59} &
\lightrightbox{0.355\linewidth}{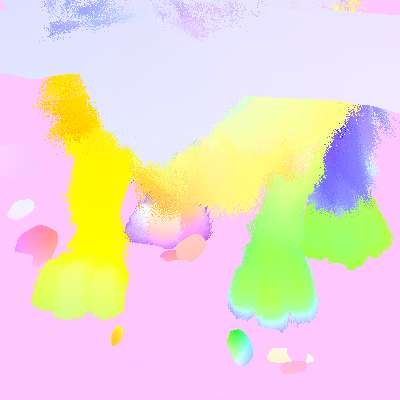}{Zoom 5.4$\times$}
\end{tabular}
\caption{Illustration of the high amount of details in the Spring dataset. The dataset consists of HD images with super-resolved UHD ground truth for disparities and optical flow.}
\label{fig:teaser}
% \vspace{-3mm}
\end{figure}

\begin{table*}[t!]
\caption{Overview over recent datasets and benchmarks (BM). Where applicable, we report available image pairs and ground truth frames for motion estimation, \ie scene flow (SF) or optical flow (OF), and for disparity estimation, \ie stereo (ST), separately.}
\label{eval:tab:overview}
\centering
\begin{adjustbox}{max width=0.98\textwidth}
\setlength\tabcolsep{4pt}
\begin{tabular}{l@{\hspace{2pt}}c@{\hspace{12pt}}c@{\hspace{5pt}}c@{\hspace{5pt}}c@{\hspace{4pt}}crr@{\hspace{10pt}}rr@{\hspace{10pt}}c@{\hspace{6pt}}c@{\hspace{6pt}}c@{\hspace{6pt}}cc}
\toprule
& Venue & SF & OF & ST & BM & \multicolumn{2}{c}{\;\;\#image pairs} & \multicolumn{2}{c}{\!\#gt\ frames} & \#pix & scenes & source & ph.realism & motion 
\\
\midrule
\textbf{Spring (ours)} & CVPR '23 & \cmark & \cmark & \cmark & \cmark & \textbf{5953} & \textbf{6000} & \textbf{23812} & \textbf{12000} & \textbf{2.1M} & \textbf{47} & \textbf{CGI} & \textbf{high} & \textbf{realistic}
\\
KITTI 2015~\cite{Menze2015_KITTI} & CVPR '15& \cmark & \cmark & \cmark & \cmark & 400 & 400 & 400 & 400 & 0.5M & n/a & real & high & automotive
\\
FlyingThings3D~\cite{Mayer2016_FTH}\! & CVPR '16 & \cmark & \cmark & \cmark & \xmark & 24084$^\ddagger$\!\!\! & 26760$^\ddagger$\!\!\! & 96336 & 53520 & 0.5M & 2676 & CGI & low & random
\\
VKITTI 2~\cite{Cabon2020_VKITTI2} & arXiv '20 & \cmark & \cmark & \cmark & \xmark & 21210 & 21260 & 84840 & 42520 & 0.5M & 5 & CGI & med. & automotive
\\
Monkaa~\cite{Mayer2016_FTH} & CVPR '16 & \cmark & \cmark & \cmark & \xmark & 8640$^\ddagger$\!\!\! & 8664$^\ddagger$\!\!\! & 34560 & 17328 & 0.5M & 8 & CGI & low & random
\\
Driving~\cite{Mayer2016_FTH} & CVPR '16 & \cmark & \cmark & \cmark & \xmark & 4392$^\ddagger$\!\!\! & 4400$^\ddagger$\!\!\! & 17568 & 8800 & 0.5M & 1 & CGI & med. & automotive
\\
\midrule
KITTI 2012~\cite{Geiger2012_KITTI} & CVPR '12 & \xmark & \cmark & \cmark & \cmark & 389 & 389 & 389 & 389 & 0.5M & n/a & real & high & automotive
\\
MPI Sintel~\cite{Butler2012_Sintel} & ECCV '12 & \xmark & \cmark & (\cmark)$^*\!\!\!$ & \cmark & 1593$^\ddagger$\!\!\! & 1064$^\ddagger$\!\!\! & 1593 & 1064 & 0.4M & 35 & CGI & high & realistic
\\
HD1K~\cite{Kondermann2016_HD1K} & CVPRW '16 & \xmark & \cmark & \xmark & \phantom{$\!\!\! ^\perp$}(\cmark)$^\dagger \!\!\!$ & 1074 & n/a & 1074 & n/a & 2.8M & 63 & real & high & automotive
\\
VIPER~\cite{Richter2017_VIPER} & ICCV '17 & \xmark & \cmark & \xmark & \cmark & 186285 & n/a & 372570 & n/a & 2.1M & 184 & CGI & high & automotive
\\
Middlebury-OF~\cite{Baker2011_MiddleburyFlow} & IJCV '11 & \xmark & \cmark & \xmark & \cmark & 16 & n/a & 16 & n/a & 0.2M & 16 & HT/CGI & med. & small
\\
Human OF~\cite{Ranjan2020_HumanMotion} & IJCV '20 & \xmark & \cmark & \xmark & \xmark & 238900 & n/a & 238900 & n/a & 0.4M & 18432 & CGI & med. & rand./human
\\
AutoFlow~\cite{Sun2021_AutoFlow} & CVPR '21 & \xmark & \cmark & \xmark & \xmark & 40000 & n/a & 40000 & n/a & 0.3M & n/a & CGI & low & random
\\
FlyingChairs~\cite{Dosovitskiy2015_FlowNet} & ICCV '15 & \xmark & \cmark & \xmark & \xmark & 22872 & n/a & 22872 & n/a & 0.2M & n/a & CGI & low & random
\\
VKITTI~\cite{Gaidon2016_VKITTI} & CVPR '16 & \xmark & \cmark & \xmark & \xmark & 21210 & n/a & 21210 & n/a & 0.5M & 5 & CGI & low & automotive
\\
Middlebury-ST~\cite{Scharstein2014_MiddleburyStereo} & GCPR '14 & \xmark & \xmark & \cmark & \cmark & n/a & 33 & n/a & 66 & 5.6M & 33 & real & high & n/a
\\
ETH3D~\cite{Schoeps2017_ETH3D} & CVPR '17 & \xmark & \xmark & \cmark & \cmark & n/a & 47 & n/a & 47 & 0.4M & 11 & real & high & n/a
\\
\bottomrule
\end{tabular}
\end{adjustbox}
\\[1mm]
{\small HT: hidden texture, $\ddagger$: available in clean and final, $*$: not part of the benchmark, $\dagger$: offline}
%\vspace{-3mm}
\end{table*}

\section{Introduction}
The estimation of dense correspondences in terms of scene flow, optical flow and disparity is the basis for numerous tasks in computer vision.
Amongst others, such tasks include action recognition, driver assistance, robot navigation, visual odometry, medical image registration, video processing, stereo reconstruction and structure-from-motion.
% Given this broad range of applications as well as their fundamental importance, it is not surprising that data-driven quantitative evaluations have ever since been an integral part of research in dense matching. 
Given this multitude of applications and their fundamental importance, datasets and benchmarks that allow %data-driven 
quantitative evaluations have ever since driven the improvement of dense matching methods. 
The introduction of suitable datasets and benchmarks did not only enable the comparison and analysis of novel methods, but also triggered the transition from classical discrete \cite{Kolmogorov2002_GraphCuts, Hirschmueller2005_SGM,Woodford2008_Stereo2ndOrder} and continuous \cite{Horn1981, Black1996, Brox2004, Huguet2007, Ranftl2014} optimization frameworks to %the 
current learning based approaches relying on neural networks \cite{Dosovitskiy2015_FlowNet,Sun2018,Teed2020_RAFT, Zhang2019_GANet, Cheng2020_LEAStereo, Teed2021_RAFT3D, Liu2022}.
%Datasets and benchmarks had a significant share in progressing the entire field of dense matching, where errors on widely-used benchmarks~\cite{Butler2012_Sintel,Menze2015_KITTI,Schoeps2017_ETH3D} are often one order of magnitude lower for state-of-the-art methods than for classical approaches.
The available benchmarks focus on distinct aspects like automotive scenarios~\cite{Geiger2012_KITTI,Menze2015_KITTI,Kondermann2016_HD1K,Richter2017_VIPER}, differing complexity of motion~\cite{Baker2007_MiddleburyFlow,Baker2011_MiddleburyFlow,Butler2012_Sintel} or
%stereo sequences with 
(un)controlled illumination~\cite{Scharstein2014_MiddleburyStereo,Schoeps2017_ETH3D}.
However, none of these benchmarks provides a combination of high-quality data and a large number of frames, to assess a method's quality in regions with fine details and to simultaneously satisfy the training needs of current neural networks. % based methods.
Furthermore, with KITTI\linebreak 2015~\cite{Menze2015_KITTI}, only a single benchmark that goes back to the pre-deep-learning era
%has been%published 
is available
for image-based scene flow, which %impedes 
currently prevents the development of well-generalizing methods due to lacking dataset variability.

\paragraphnew{Contributions.} 
To tackle these challenges, we propose the \emph{Spring dataset and benchmark}, providing a large number of high-quality and high-resolution frames and ground truths to enable the development of even more accurate methods for scene flow, optical flow and stereo estimation.
% With Spring, we complement existing benchmarks through a focus on high-detail data, which directly supports the development of methods that generalize across datasets with varying properties.
With Spring, we complement existing benchmarks through a focus on high-detail data, while we simultaneously broaden the number of available datasets for the development of well-generalizing methods across data with varying properties. %; see e.g.\ the Robust Vision Challenge\footnote{{\tt http://www.robustvision.net/}}.
The latter aspect is particularly valuable for image-based scene flow methods.
There, we provide the first benchmark with high-resolution, dense ground truth data in the literature.
In summary, our contributions are fourfold: 
 \begin{enumerate}[(i)]
 \setlength{\itemsep}{0\baselineskip}
 \item 
% (i)
{\em New dataset:} Based on the open-source Blender movie \enquote{Spring}, we rendered 6000 stereo image pairs from 47 sequences with state-of-the-art visual effects in HD resolution (1920$\times$1080px). For those image pairs, we extracted ground truth from Blender in forward and backward direction, both in space and time, %scene flow, optical flow and disparity ground truth from Blender, 
amounting to 12000 ground truth frames for stereo and 23812 ground truth frames for motion -- 60$\times$ more than KITTI and 15$\times$ more than MPI Sintel.
\item
{\em High-detail evaluation methodology:} To adequately assess small details at a pixel level, %such as grass or hair,
we propose a novel evaluation methodology that relies on an even higher resolved ground truth. All %35812 
ground truth frames are computed in UHD resolution (3840$\times$2160px). 
\item
{\em Benchmark:} We set up a public benchmark website to upload, analyze and compare novel methods. 
It provides several widely used error measures and additionally analyzes the results in different types of regions, including high-detail, unmatched, non-rigid, sky and large-displacement areas.
\item
{\em Baselines:} We evaluated 15 state-of-the-art methods\linebreak (8 optical flow, 4 stereo, 3 scene flow) as non-fine\-tuned baselines.
Results not only show that small details still pose a problem to recent methods, but also hint at significant potential improvements in all tasks.
\end{enumerate}

\begin{figure*}
\centering
% \setlength\tabcolsep{2pt}
% \def\arraystretch{0.0}
% \begin{tabular}{ccc}
\darklefttbox{0.32\textwidth}{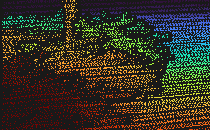}{KITTI 15} \hspace{1mm}
\darklefttbox{0.32\textwidth}{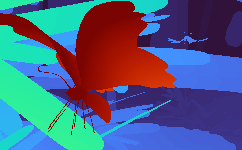}{MPI Sintel} \hspace{1mm}
\darklefttbox{0.32\textwidth}{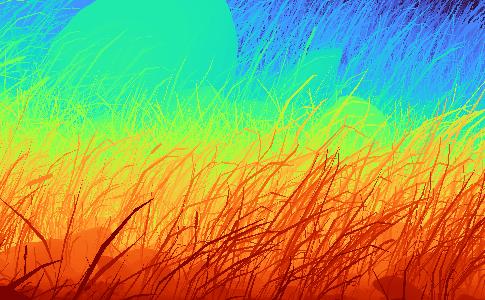}{Spring (ours)}
% \end{tabular}
\caption{Ground truth comparison for highly-detailed regions across datasets. \emph{From left to right}: KITTI 15~\cite{Menze2015_KITTI}, MPI Sintel~\cite{Butler2012_Sintel}, Spring.}
\label{fig:details}
\end{figure*}

\section{Related work}
In the literature there exists a large number of datasets and benchmarks for dense matching covering scene flow, optical flow and stereo. An overview over the state of the art is given in \cref{eval:tab:overview}.
Datasets and benchmarks can roughly be divided along two orthogonal axes: The scene axis and the data axis.
On one edge of the scene axis, there are datasets focusing on \emph{automotive scenes} such as KITTI~\cite{Geiger2012_KITTI,Menze2015_KITTI}, Virtual KITTI~\cite{Gaidon2016_VKITTI,Cabon2020_VKITTI2}, HD1K~\cite{Kondermann2016_HD1K}, VIPER~\cite{Richter2017_VIPER} and Driving~\cite{Mayer2016_FTH}.
On the other edge are datasets that target \emph{general scenes} such as Middlebury~\cite{Baker2007_MiddleburyFlow,Scharstein2014_MiddleburyStereo}, MPI Sintel~\cite{Butler2012_Sintel}, ETH3D~\cite{Schoeps2017_ETH3D}, FlyingThings3D~\cite{Mayer2016_FTH}, Monkaa~\cite{Mayer2016_FTH} and FlyingChairs~\cite{Dosovitskiy2015_FlowNet} with the special case of Human OF \cite{Ranjan2020_HumanMotion} that addresses human body motion.
Along the orthogonal data axis, datasets and benchmarks can be divided into \emph{real-world data}~\cite{Geiger2012_KITTI,Menze2015_KITTI,Kondermann2016_HD1K,Scharstein2014_MiddleburyStereo,Schoeps2017_ETH3D} and \emph{synthetic data}~\cite{Butler2012_Sintel,Richter2017_VIPER,Mayer2016_FTH,Dosovitskiy2015_FlowNet,Gaidon2016_VKITTI,Ranjan2020_HumanMotion,Sun2021_AutoFlow}.
In the context of optical flow, Sintel showed that synthetic benchmarks can validly approximate the statistics of natural images and motion, while avoiding ground truth measurements in the real world.
This motivates the use of synthetic data for general scenes in our Spring dataset. % and benchmark.

\paragraphnew{Resolution.}
Most of the benchmarks and datasets are limited to images with QHD resolution (0.5 Mpix) or below~\cite{Geiger2012_KITTI, Menze2015_KITTI, Butler2012_Sintel, Mayer2016_FTH, Dosovitskiy2015_FlowNet, Schoeps2017_ETH3D, Ranjan2020_HumanMotion, Sun2021_AutoFlow, Greff2022_Kubric}.
Only in the case of automotive optical flow, full HD resolution (2-3 Mpix) has been considered~\cite{Kondermann2016_HD1K, Richter2017_VIPER}.
Moreover, first attempts with image sizes beyond HD (5.6 Mpix) have been made for real-world stereo, but only for a small number of images~\cite{Scharstein2014_MiddleburyStereo}.
With respect to general optical flow and scene flow, Spring is the first dataset to consider \emph{HD resolution and above} for input images and ground truth, respectively.

\paragraphnew{Dataset size.}
Considering the \emph{number of input and ground truth images}, most popular benchmarks have at most 1600 frames.
%are limited to roughly 1500 frames or below.
This is either due to the use of real-world footage~\cite{Geiger2012_KITTI,Scharstein2014_MiddleburyStereo,Menze2015_KITTI,Kondermann2016_HD1K, Schoeps2017_ETH3D} or their creation in the pre-deep-learning era~\cite{Geiger2012_KITTI,Butler2012_Sintel,Menze2015_KITTI}. 
The only exception is the automotive  optical flow benchmark VIPER~\cite{Richter2017_VIPER}, if we leave aside several large training datasets without benchmark functionality such as FlyingChairs~\cite{Dosovitskiy2015_FlowNet}, Virtual KITTI~\cite{Gaidon2016_VKITTI,Cabon2020_VKITTI2}, FlyingThings3D with Monkaa and Driving~\cite{Mayer2016_FTH}, Human OF~\cite{Ranjan2020_HumanMotion}, AutoFlow \cite{Sun2021_AutoFlow} and the dataset generator Kubric~ \cite{Greff2022_Kubric}.
With respect to the number of frames for stereo and scene flow, however, Spring is the first general-scene benchmark with \emph{several thousand samples} for training and testing, closing 
a previously existing gap in the available training data for deep-learning matching methods for those tasks.

\paragraphnew{Benchmark evaluation.}
Regarding \emph{focused evaluations}
in particularly important regions, current benchmarks mainly focus on occlusions (unmatched regions)~\cite{Baker2011_MiddleburyFlow,Butler2012_Sintel, Menze2015_KITTI,Scharstein2014_MiddleburyStereo,Schoeps2017_ETH3D}, discontinuities~\cite{Baker2007_MiddleburyFlow,Butler2012_Sintel}, large displacements~\cite{Butler2012_Sintel} and non-rigid areas~\cite{Menze2015_KITTI}.
%\textcolor{red}{Sollte hier auch noch was zu rigid/non-rigid stehen? Nochmal checken dass hier alle evaluations-maps gelistet werden, und unten im letzten Satz richtig genannt werden.}
However, recent stereo~\cite{Li2022_CREStereo} and optical flow methods~\cite{Jahedi2022_MSRAFT,Jahedi2022_MSRAFT_RVC} achieve highly detailed results due to cascaded recurrent neural networks that process images at larger resolutions and omit a significant upsampling of the final results (e.g.\ in contrast to~\cite{Teed2020_RAFT,Jiang2021_GMA,Lipson2021_RAFTStereo}).
This raises the question how well such high-quality methods can estimate fine-scale details such as grass or hair.
In this context, Spring not only provides high-resolution images that contain \emph{small-scale details} at pixel level, but also provides a novel \emph{evaluation methodology} that allows to  measure the accuracy in the presence of thin structures.
Therefore, the Spring benchmark provides \emph{focused evaluations} for regions with unmatched, high-detail, non-rigid,
%and sky pixels.
sky and large-displacement pixels.

\paragraphnew{Scene flow benchmarking.}
Finally, with only 400 frames, %for training and evaluation, respectively,
KITTI 2015~\cite{Menze2015_KITTI} is the only benchmark that, besides optical flow and stereo, also allows the evaluation of scene flow. Moreover, while there are a few datasets and challenges for scene flow from RGB-D \cite{Sturm2012_RGBD,Shao2018_RGBD,Lv2018_RGBD} and LiDAR \cite{Sun2020_Waymo} data, KITTI 2015 is the only benchmark for image-based scene flow, \ie scene flow only from stereo image pairs. As a consequence, recent image-based scene flow methods perform well on KITTI 2015 but are likely not robust under other types of data.
As shown for optical flow and stereo in context of the Robust Vision Challenge \cite{rvc}, improving the generalization across benchmarks is essential to increase applicability and robustness~\cite{Schmalfuss2022}.
% towards general applicability.
%generalization across benchmarks is hard to achieve. 
Hence, having \emph{more specifically tailored benchmarks} at hand %that allow the evaluation of scene flow, 
%in particular in a non-automotive setting, 
would not only be beneficial for optical flow and stereo, but in particular for image-based scene flow, where
they are \emph{indispensable} to further advance research. 
While with FlyingThings3D, Monkaa and Driving \cite{Mayer2016_FTH} as well as VKITTI~2 \cite{Cabon2020_VKITTI2} there are a few datasets available that could also be used for benchmarking, these datasets provide the ground truth for all frames which %in turn 
encourages overfitting. %Hence, 
In contrast, Spring offers the {\em full benchmark functionality}, \ie hidden ground truth for the test set, an evaluation protocol, cheating prevention and a website to compare results.

\section{Spring dataset}
\label{sec:dataset}
The Spring dataset is a novel, large, computer-generated dataset for training and evaluating scene flow, optical flow and stereo methods.
Our dataset consists of stereoscopic video sequences and ground truth scene flow in its standard parametrization with reference frame disparity, 
%(\emph{disp1}), 
target frame disparity %(\emph{disp2}) 
and optical flow \cite{Huguet2007}.
We provide ground truth data for all available combinations; for the left and right view as well as motion in temporal forward and backward direction.
The Spring dataset is based on the %Blender 
open-source movie \enquote{Spring}, which we utilize to generate a large dataset that is suitable for motion and disparity estimation.
In the following, we introduce the underlying movie data, give details on the dataset creation process and provide a full overview of our data along with a comparison to other established datasets.

\subsection{Open-source movie data}
%license
We retrieve our data from an open movie project generated in the open-source 3D software Blender.
The core idea of open movies is to make all assets that are required to render the movie available under an open license, which enables anybody to build upon the creative work and to utilize it for research~\cite{Butler2012_Sintel,Mayer2016_FTH,Im_2022_ECCV}.
The \enquote{Spring} movie is a recent project showcasing the current progress in Blender 2.80 and the ray tracing engine Cycles and is with its scenes and assets available under the open CC BY 4.0 license.
The movie scenes depict a large range of shot sizes from extreme close-ups to very long shots and a large range of motions from animated creatures, flight sequences, chasing motion, plant growth, and physically plausible simulations of pebbles, grass and hair motion.
It shows advanced visual effects that work towards a realistic appearance of the computer-generated data, such as 3D motion blur and camera depth of field with focus pulls (changes of the focal plane).
Some scenes even contain zooms (changes of the focal length, \ie frame-dependent camera intrinsics), which makes our dataset the first scene flow dataset~\cite{Menze2015_KITTI,Mayer2016_FTH,Cabon2020_VKITTI2} with this property.
In general, all movie assets are highly detailed, see \eg \cref{fig:teaser,fig:details} which provides a good basis for 
% the creation of
our datasets.
% For each step, more details are given below.

\begin{figure}
\centering
\setlength\tabcolsep{1pt}
\setlength{\fboxrule}{.75pt}%
\setlength{\fboxsep}{0pt}%
\def\arraystretch{0.0}
\begin{tabular}{@{}cc@{}}
\mapdetails{0.495\linewidth}{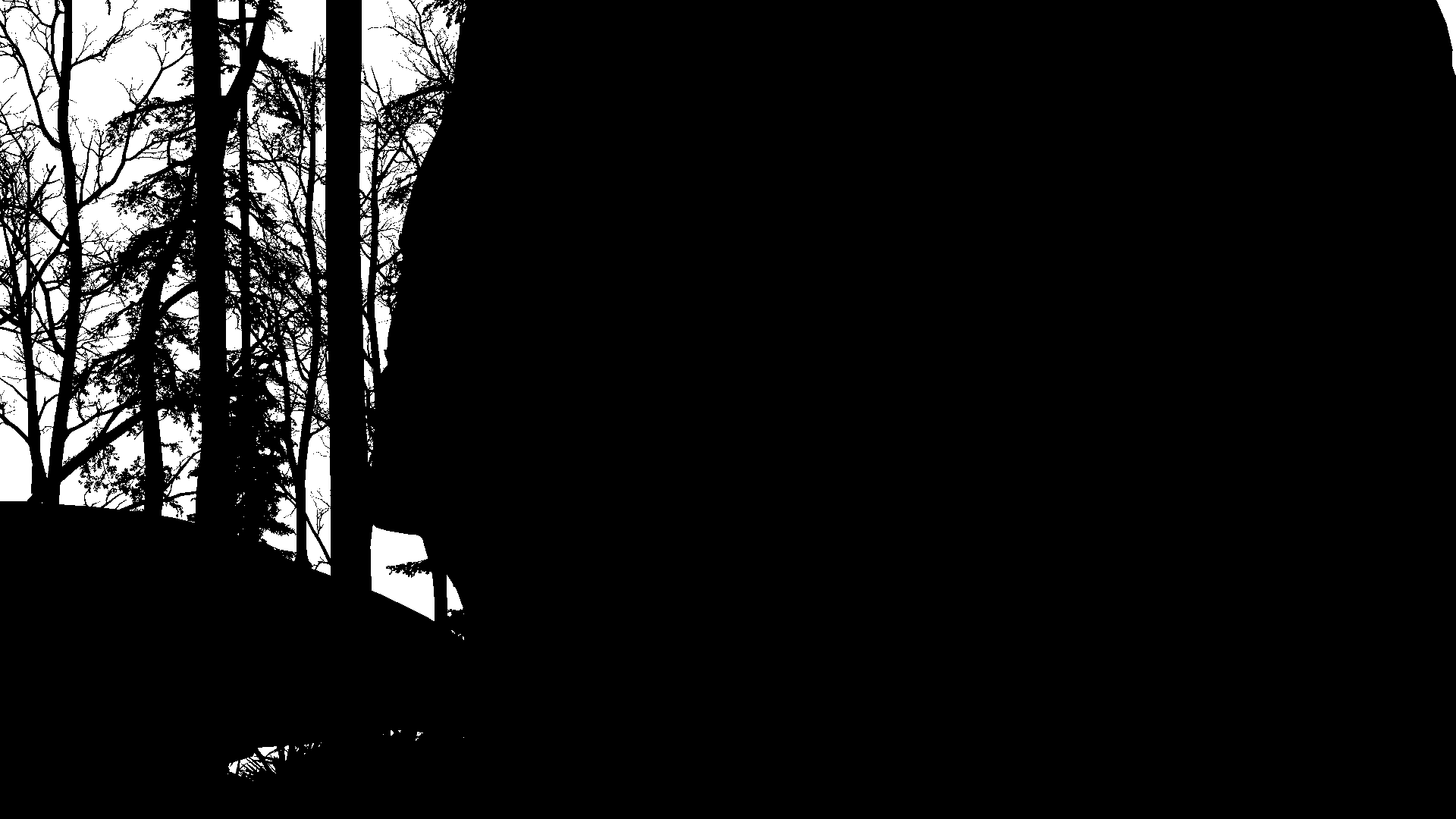}{0.155\linewidth}{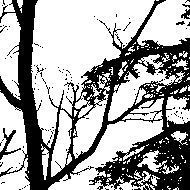}{-1.95}{0.56}{Sky map} & 
\mapdetails{0.495\linewidth}{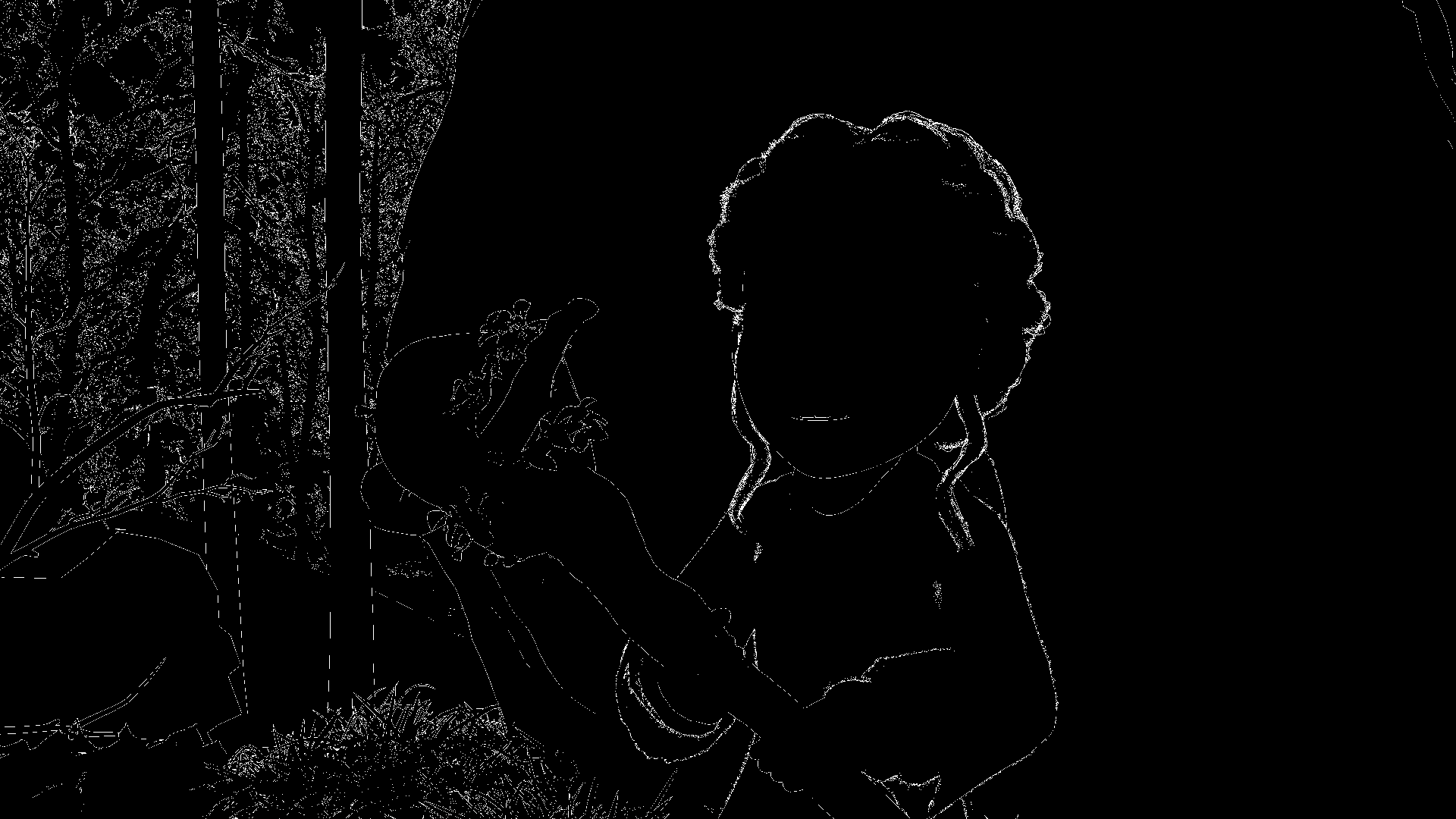}{0.155\linewidth}{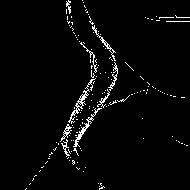}{-.1}{-0.4}{Detail map} \\[2pt]
\mapdetails{0.495\linewidth}{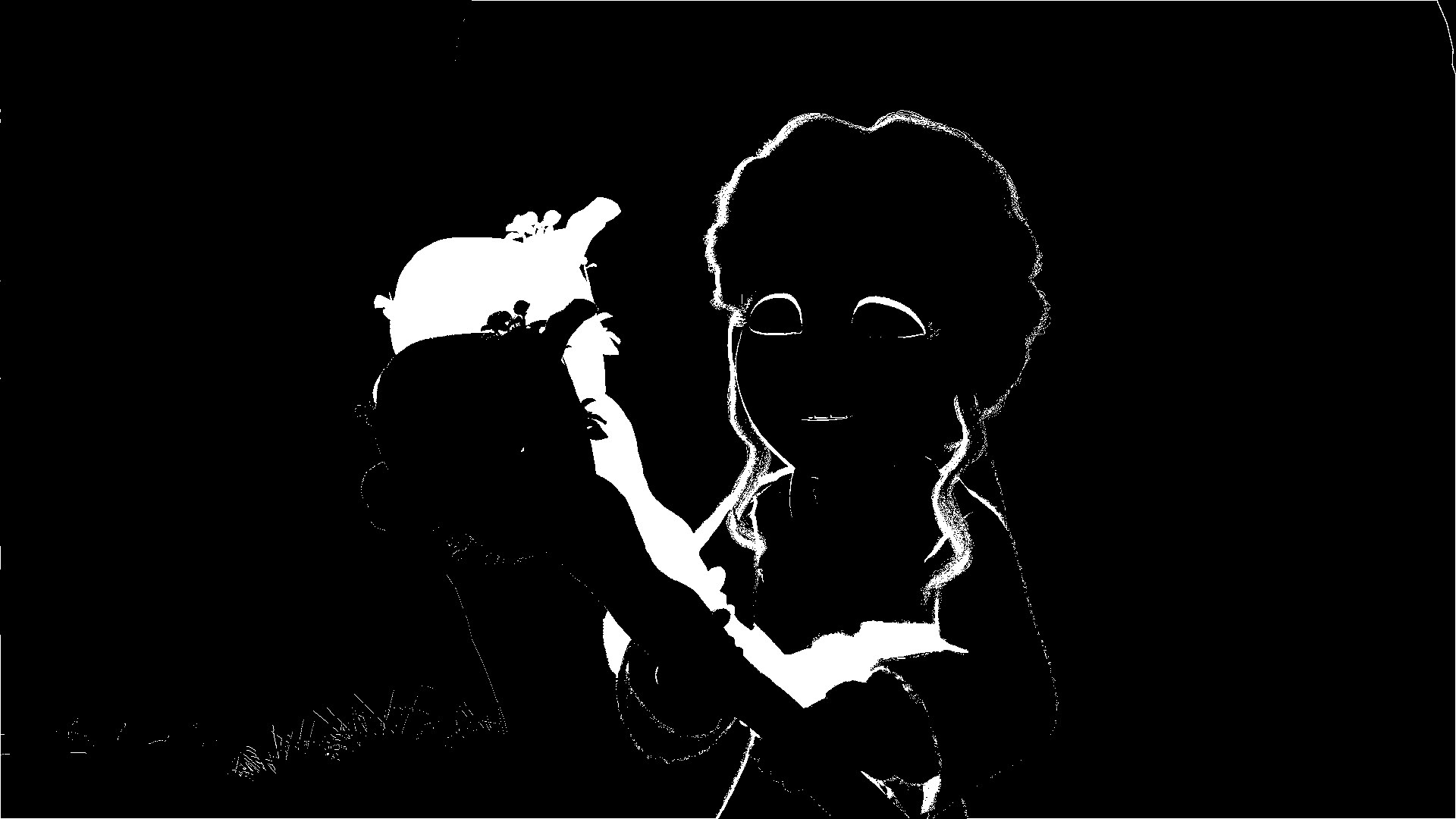}{0.155\linewidth}{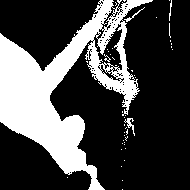}{-.18}{-.62}{Unmatched regions} & 
\mapdetails{0.495\linewidth}{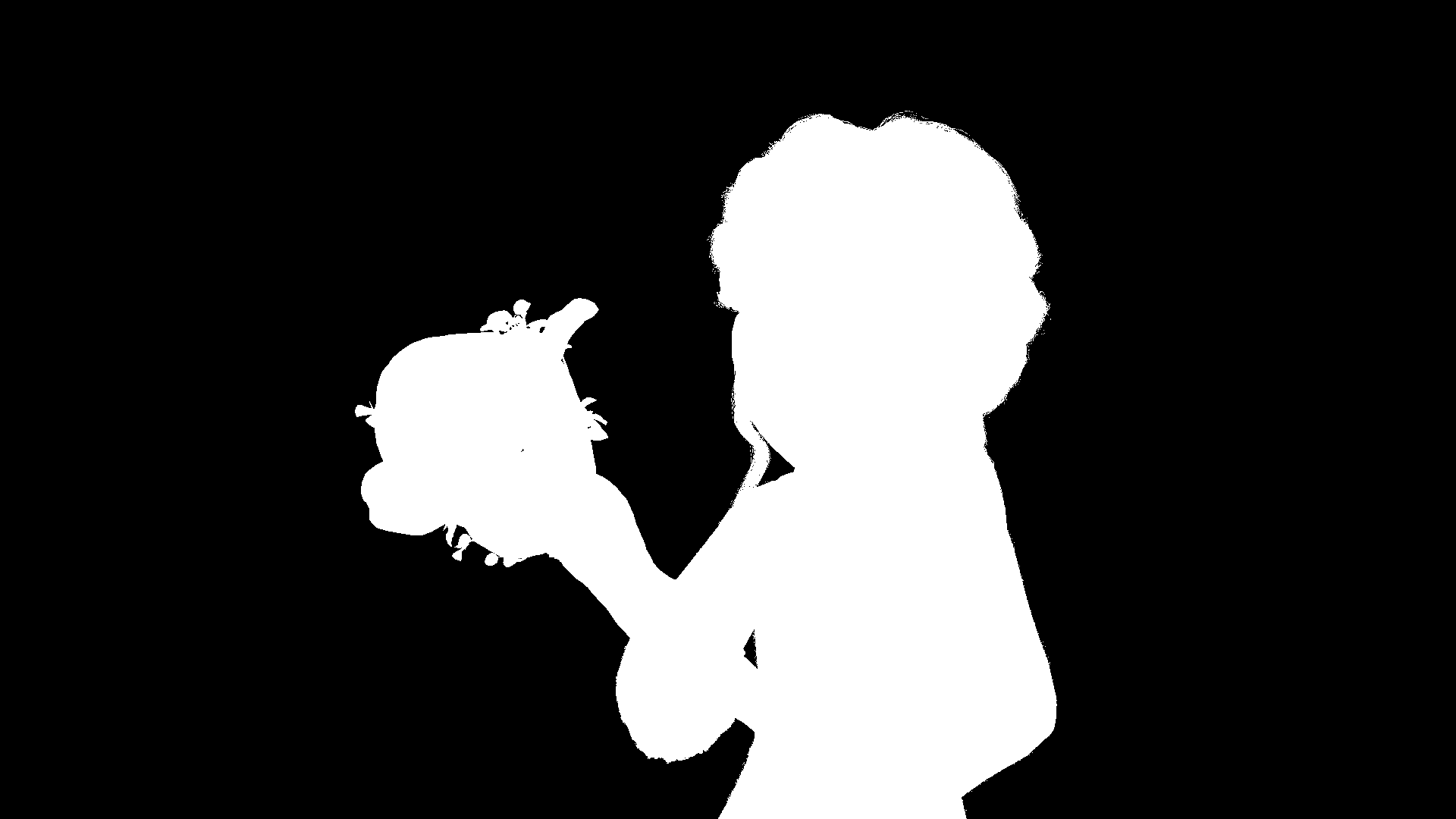}{0.155\linewidth}{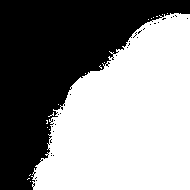}{-.11}{0.45}{Non-rigid motion}
\end{tabular}
\caption{Examples for evaluation maps. \emph{Top row}: Sky map, detail map, \emph{bottom row}: unmatched regions map, non-rigid motion map.}
\label{fig:maps}
\end{figure}

% \iffalse
\begin{figure*}
\centering
{
\setlength\tabcolsep{1.5pt}
\def\arraystretch{0.0}
\newcommand*\tabimsize{0.160}
\begin{tabular}{@{}cccccccc@{}}
\splitimg{\tabimsize\linewidth}{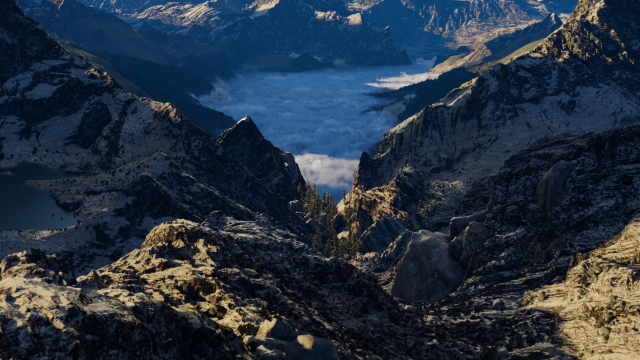}{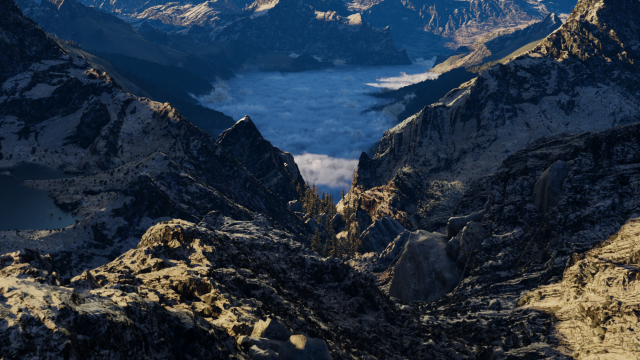}&
\splitimg{\tabimsize\linewidth}{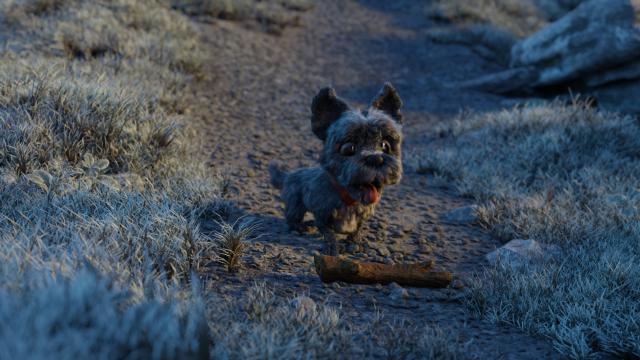}{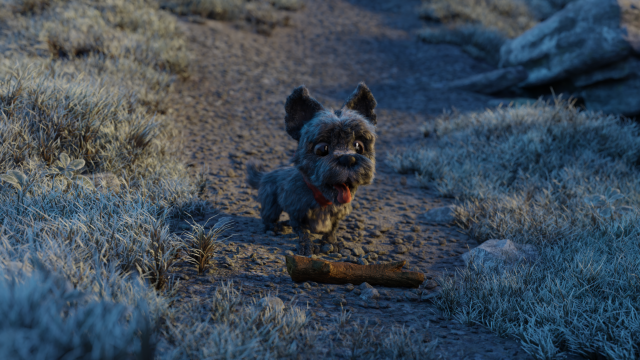}&
\splitimg{\tabimsize\linewidth}{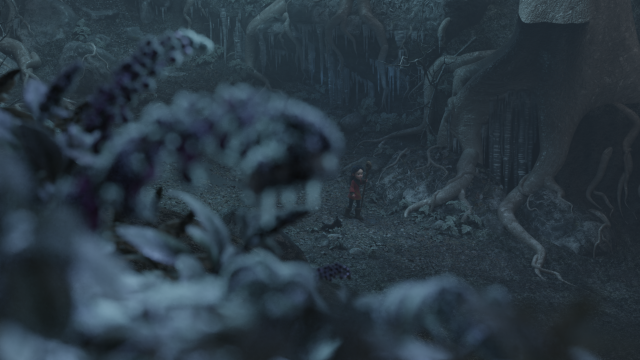}{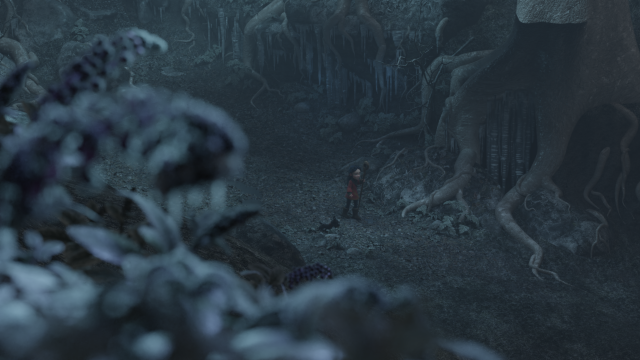}&
\splitimg{\tabimsize\linewidth}{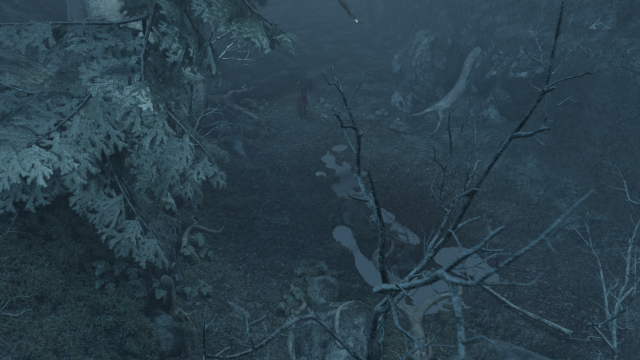}{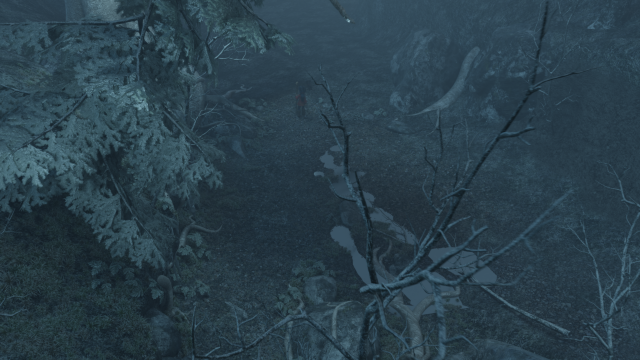}&
\splitimg{\tabimsize\linewidth}{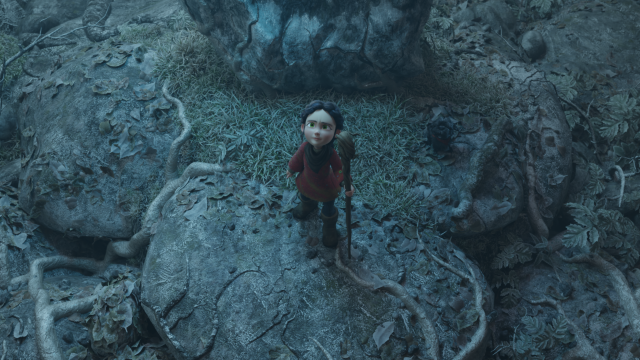}{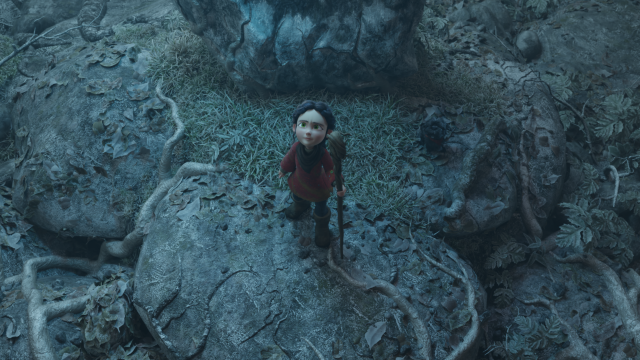}&
\splitimg{\tabimsize\linewidth}{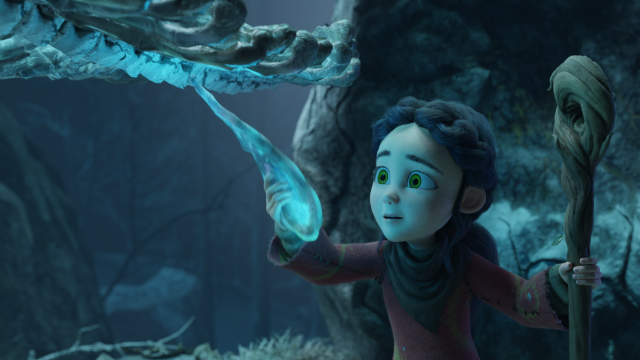}{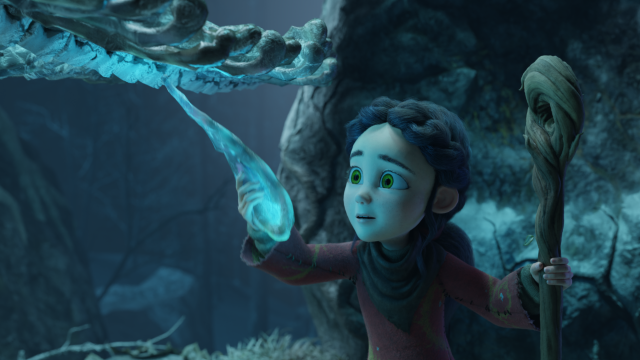}
\\[1pt]
\splitimg{\tabimsize\linewidth}{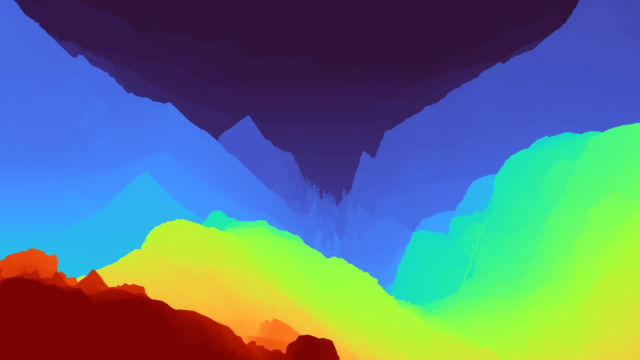}{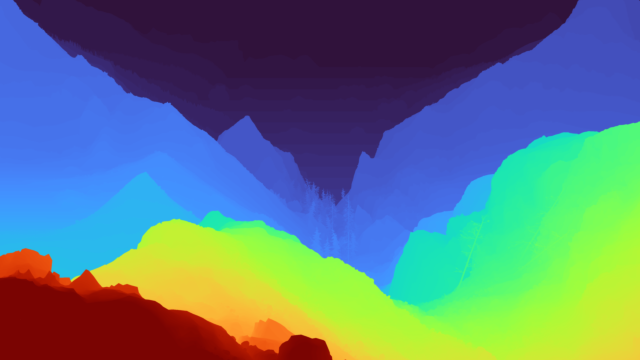}&
\splitimg{\tabimsize\linewidth}{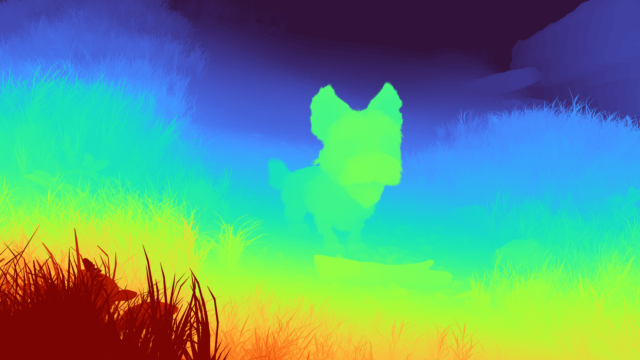}{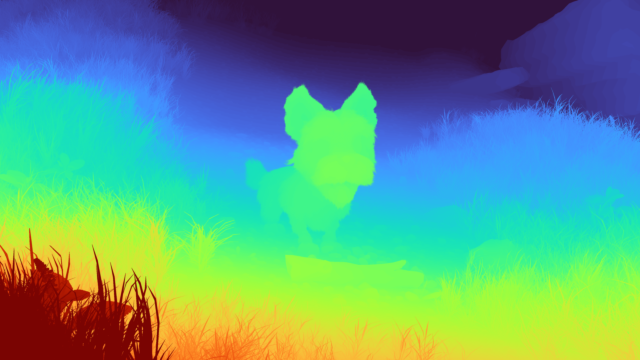}&
\splitimg{\tabimsize\linewidth}{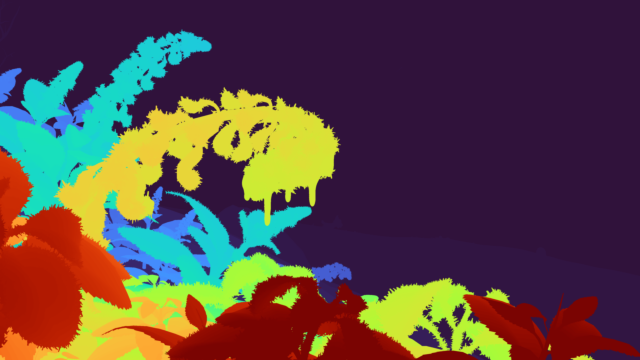}{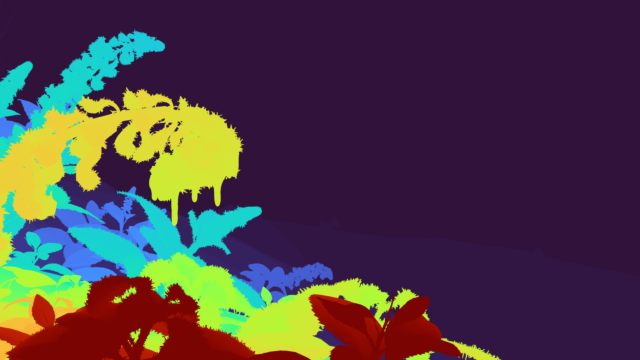}&
\splitimg{\tabimsize\linewidth}{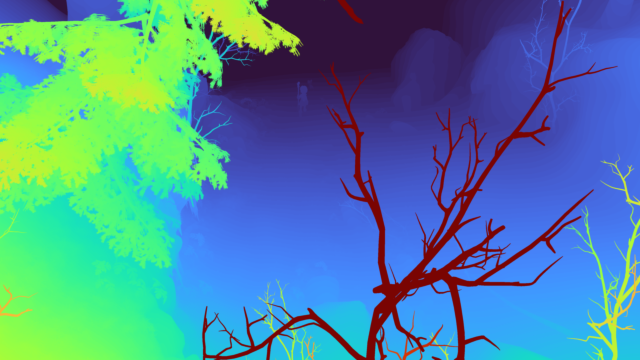}{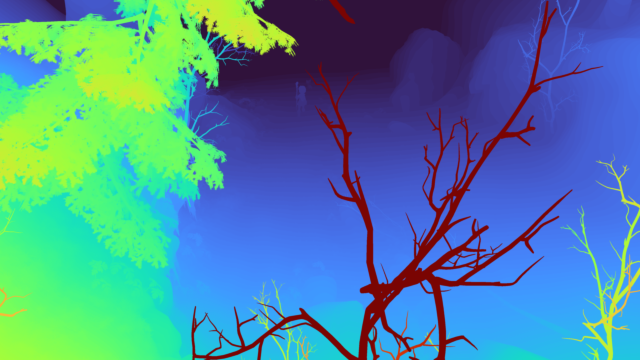}&
\splitimg{\tabimsize\linewidth}{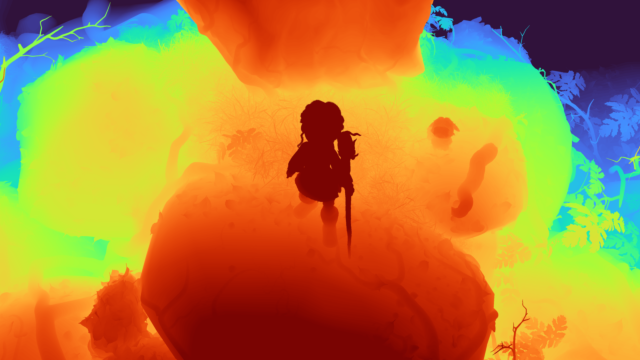}{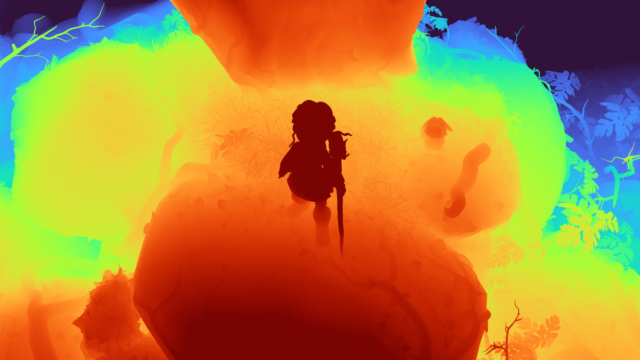}&
\splitimg{\tabimsize\linewidth}{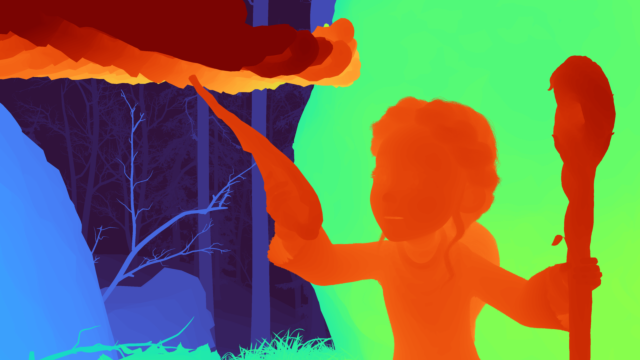}{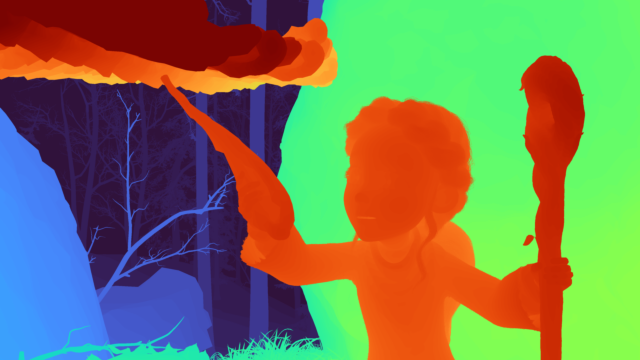}
\\[1pt]
\splitimgfour{\tabimsize\linewidth}{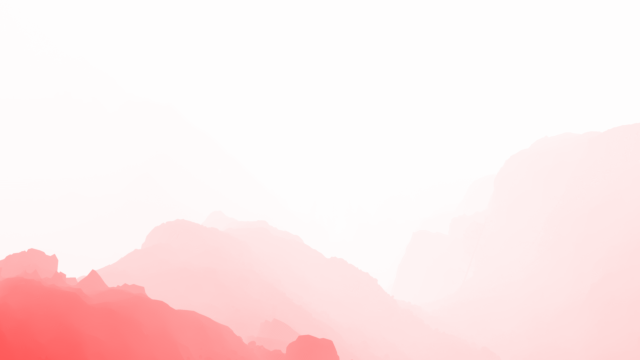}{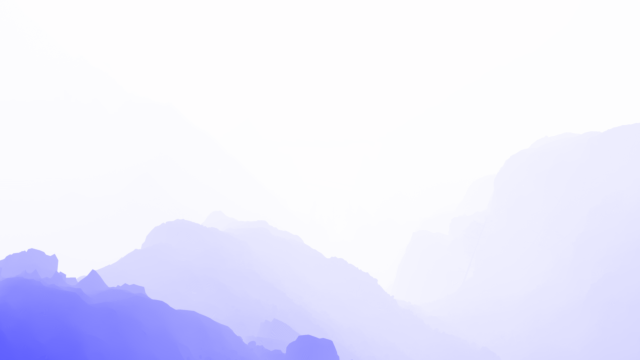}{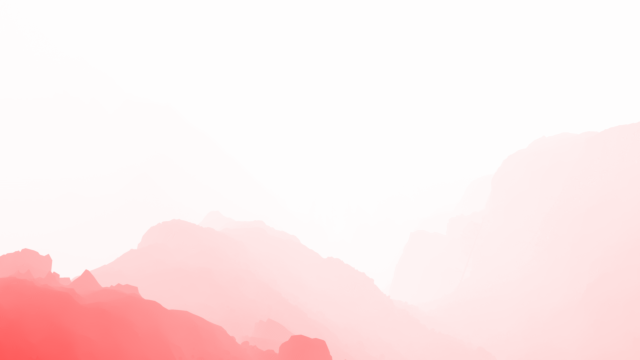}{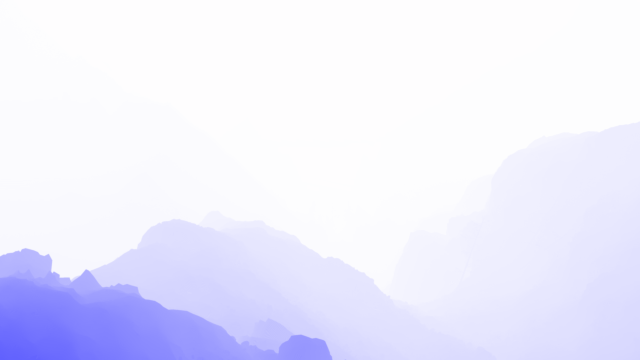}&
\splitimgfour{\tabimsize\linewidth}{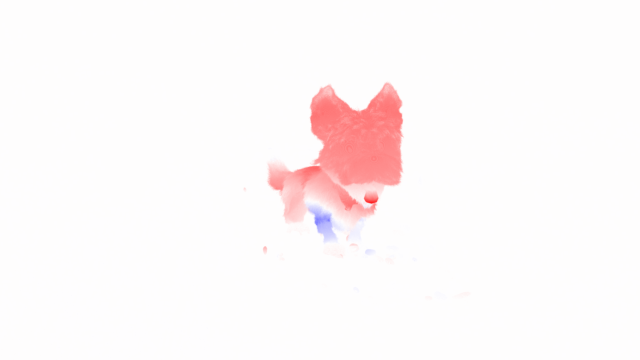}{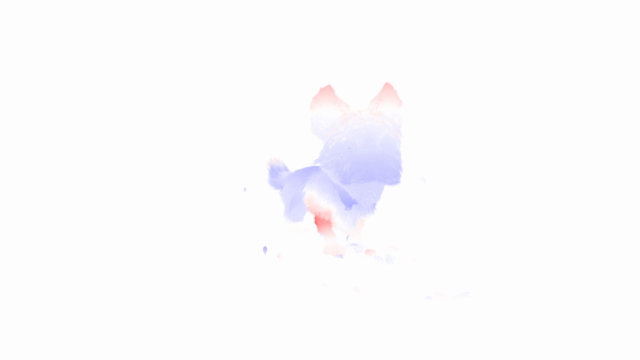}{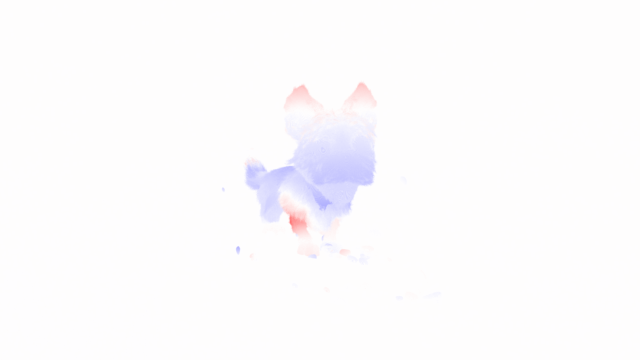}{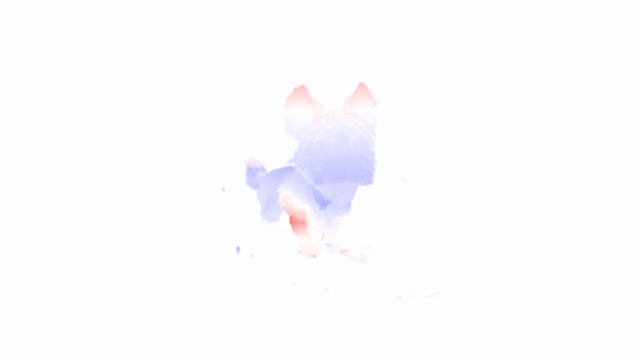}&
\splitimgfour{\tabimsize\linewidth}{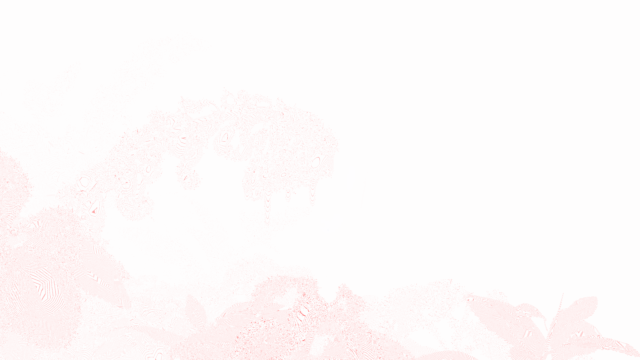}{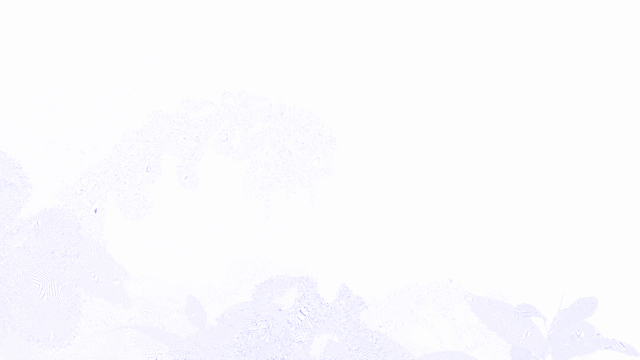}{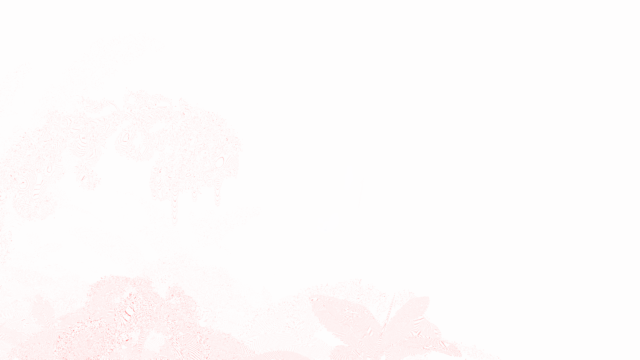}{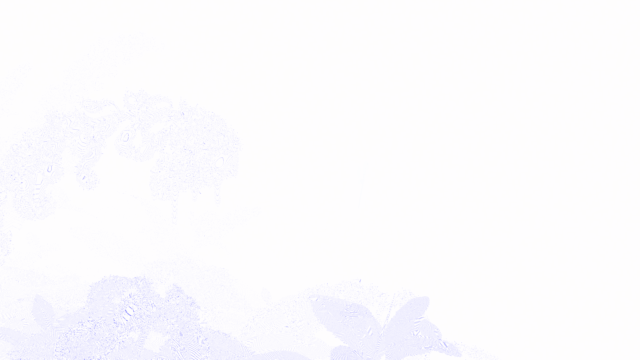}&
\splitimgfour{\tabimsize\linewidth}{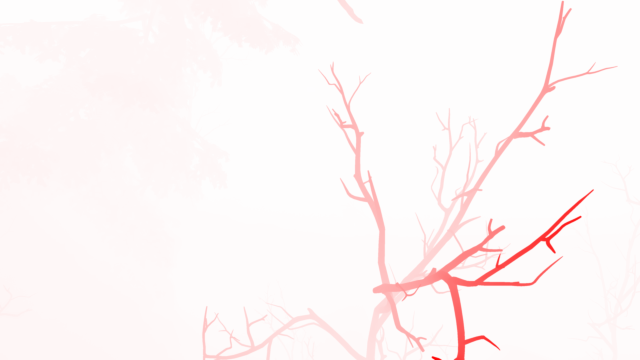}{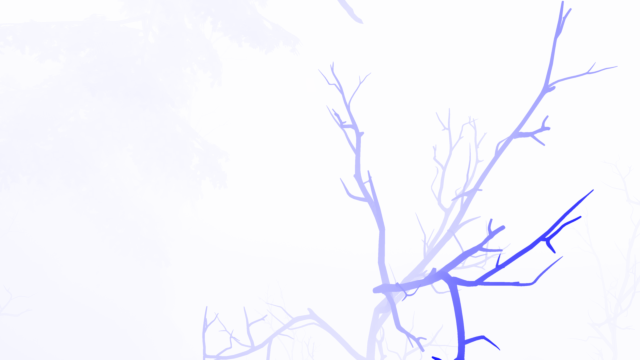}{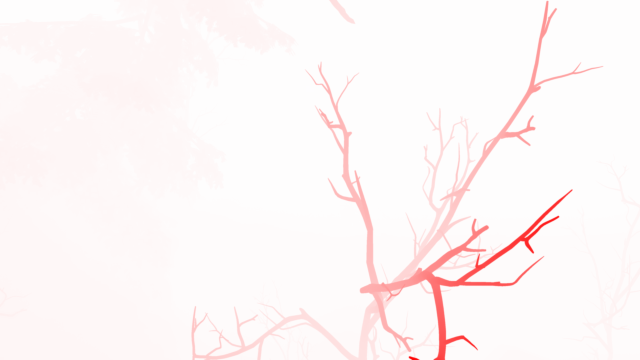}{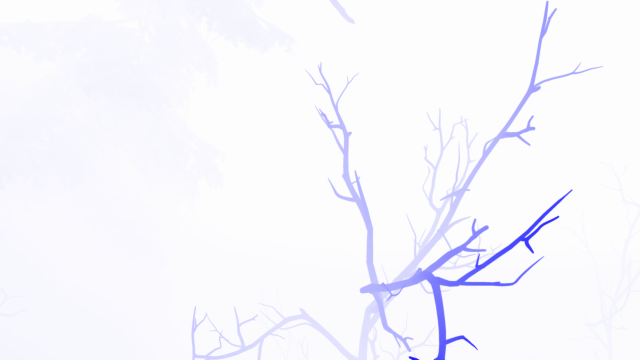}&
\splitimgfour{\tabimsize\linewidth}{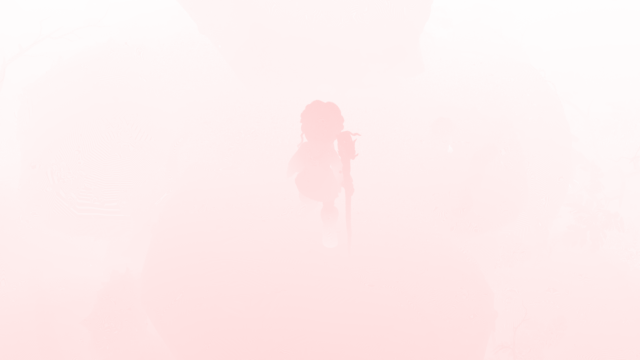}{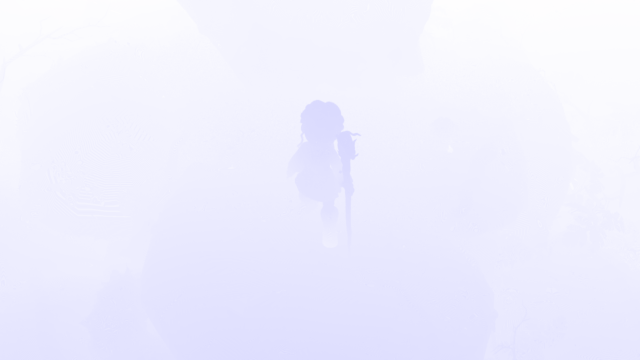}{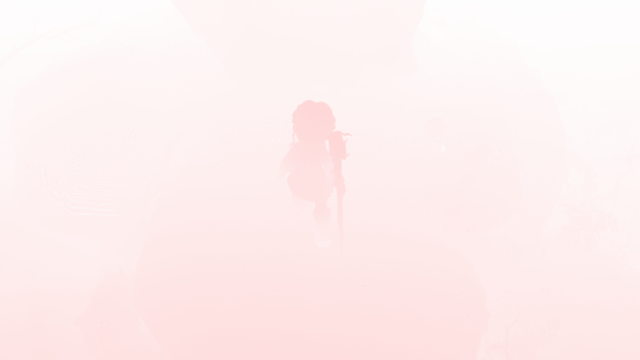}{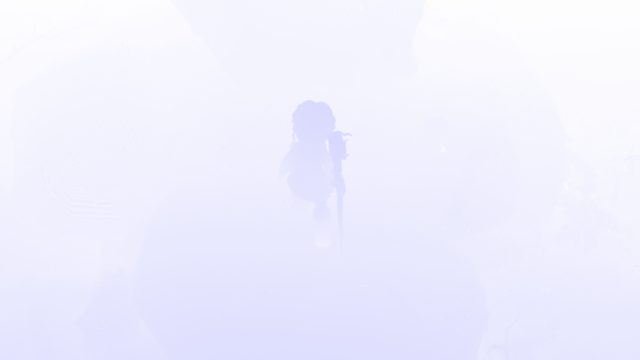}&
\splitimgfour{\tabimsize\linewidth}{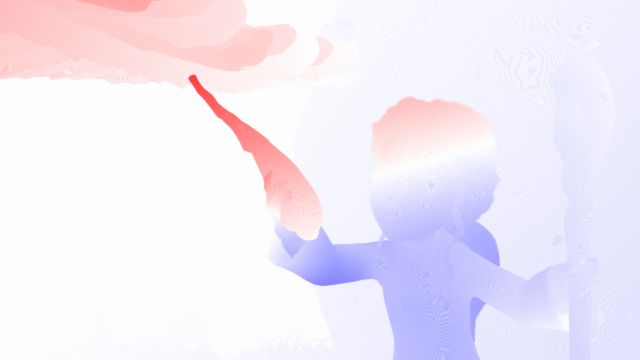}{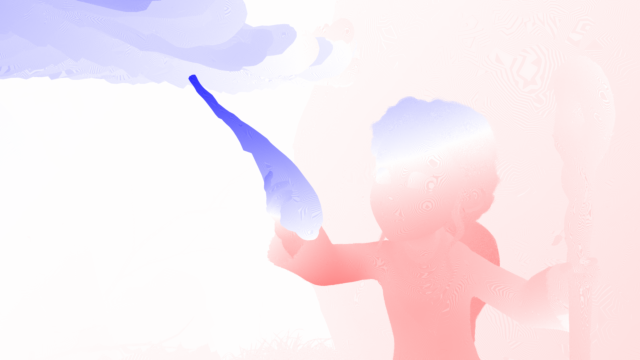}{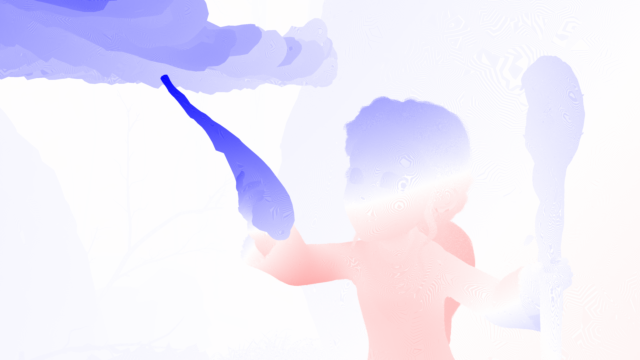}{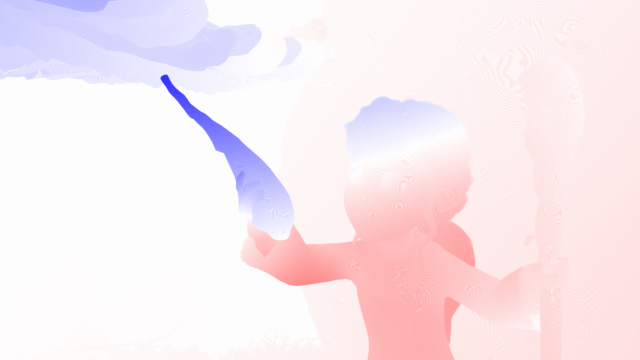}
\\[1pt]
\splitimgfour{\tabimsize\linewidth}{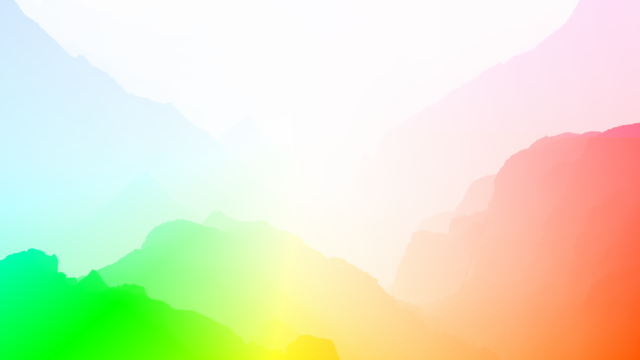}{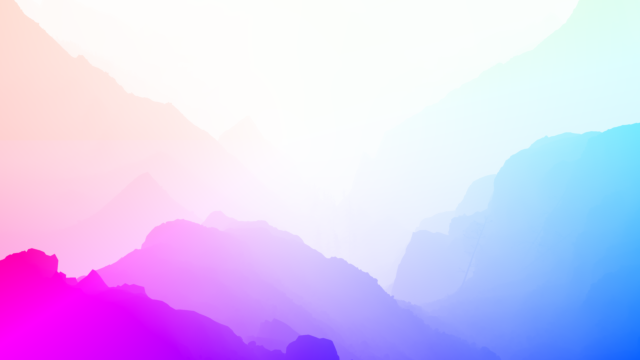}{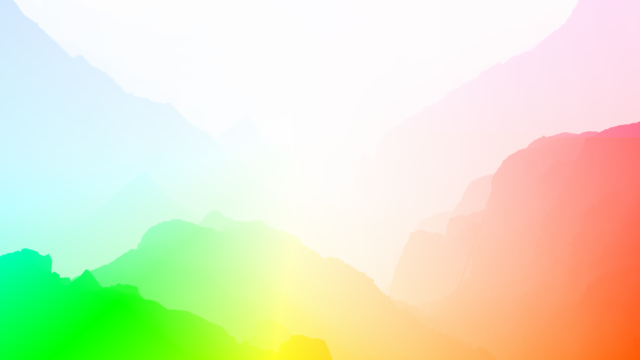}{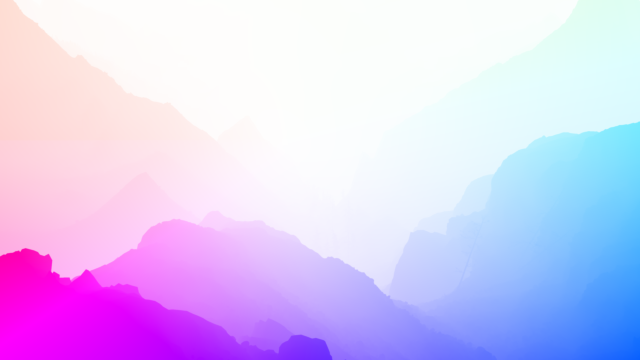}&
\splitimgfour{\tabimsize\linewidth}{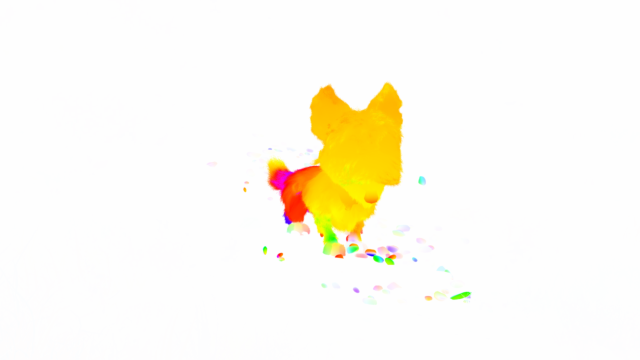}{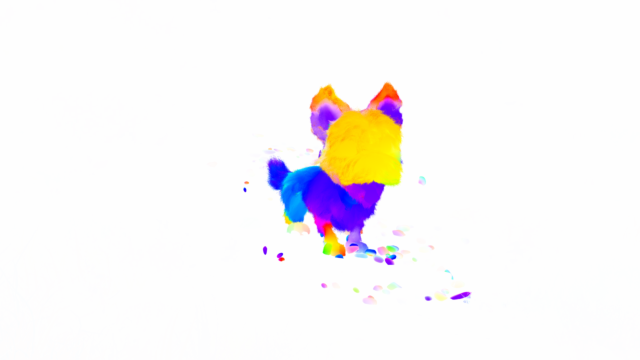}{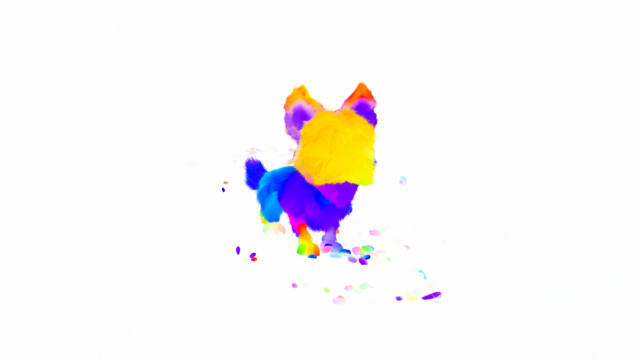}{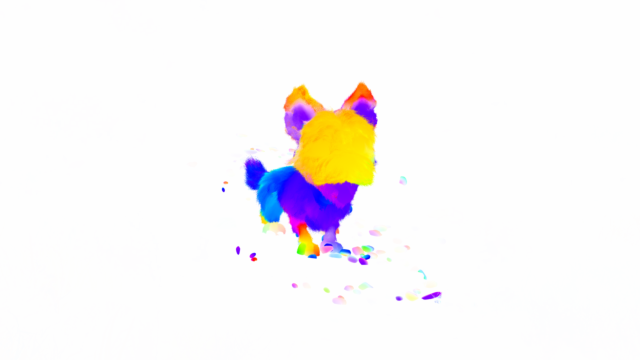}&
\splitimgfour{\tabimsize\linewidth}{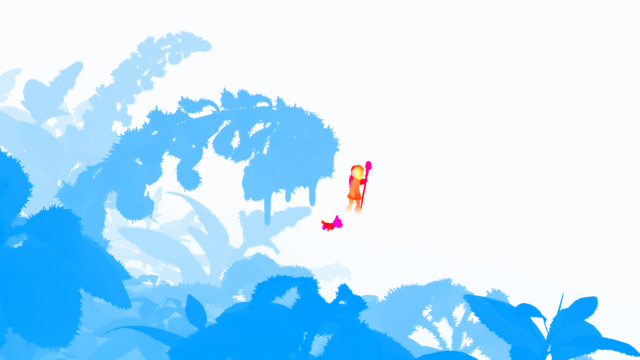}{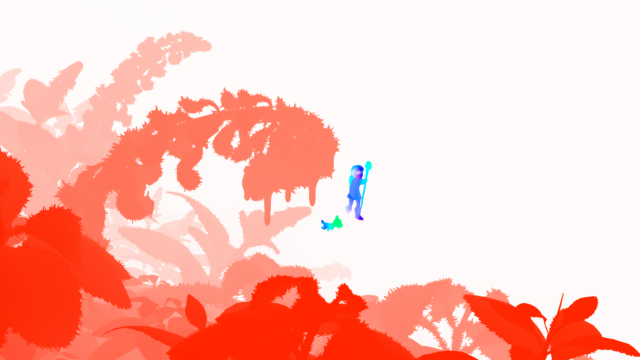}{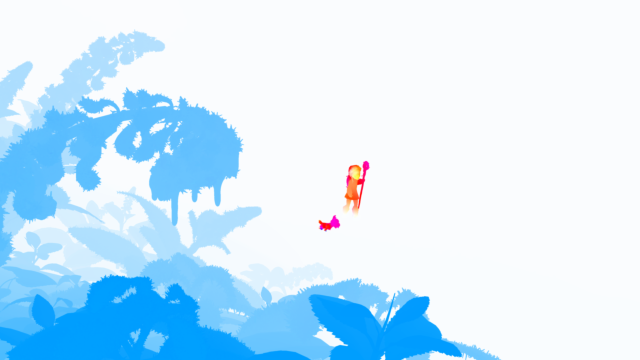}{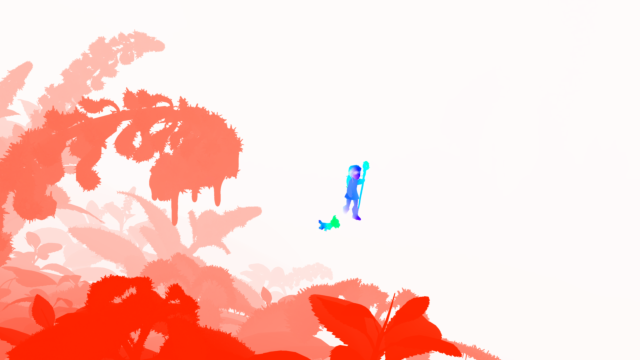}&
\splitimgfour{\tabimsize\linewidth}{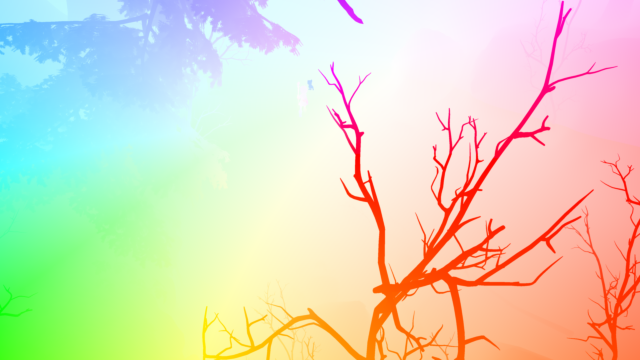}{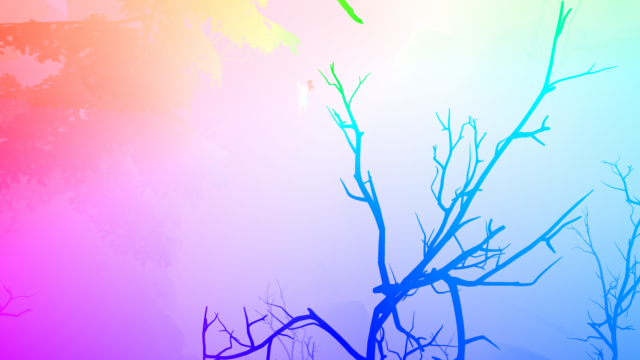}{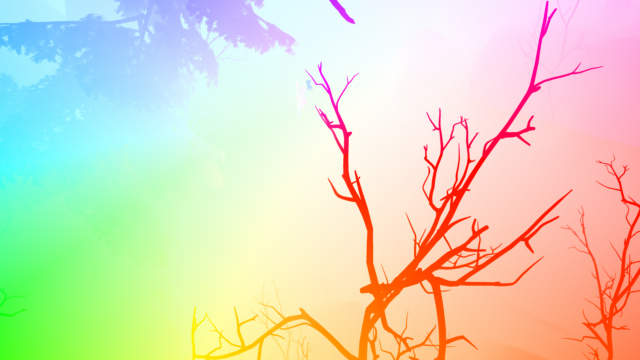}{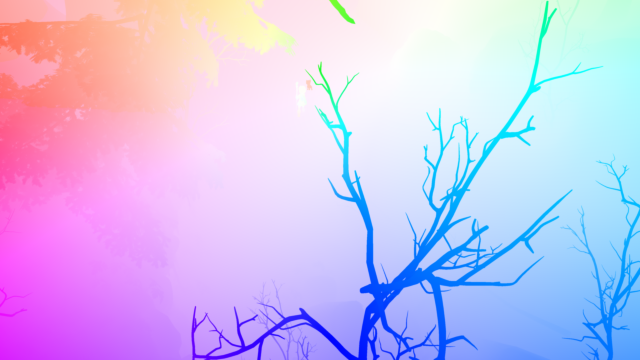}&
\splitimgfour{\tabimsize\linewidth}{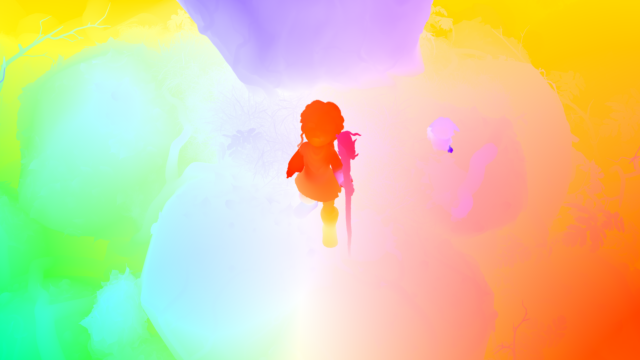}{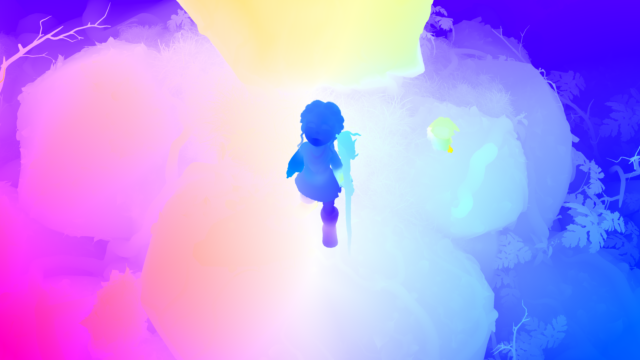}{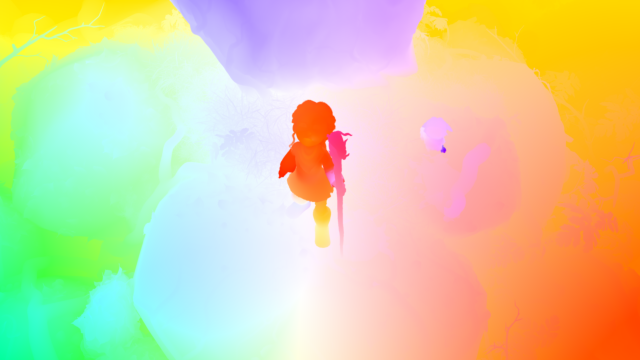}{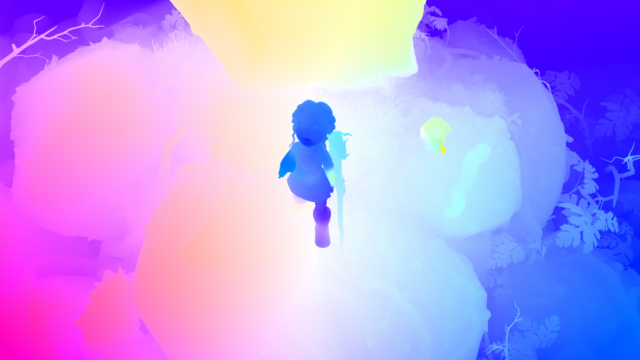}&
\splitimgfour{\tabimsize\linewidth}{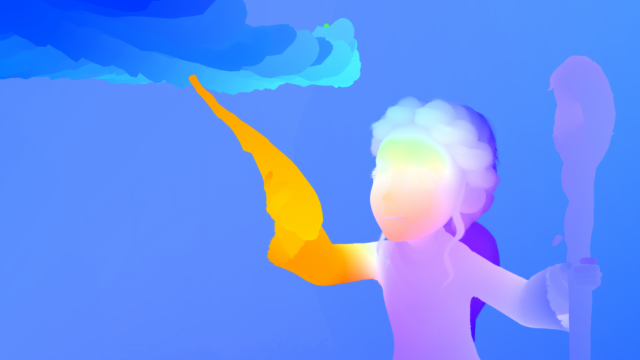}{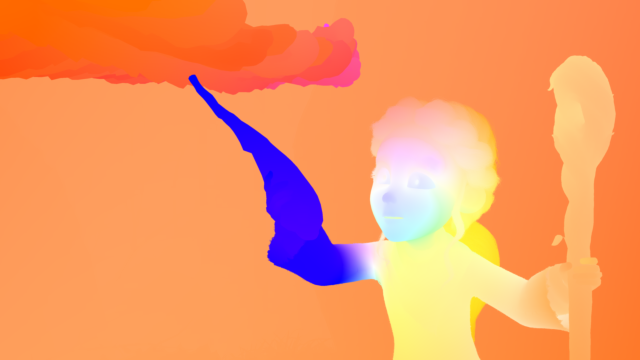}{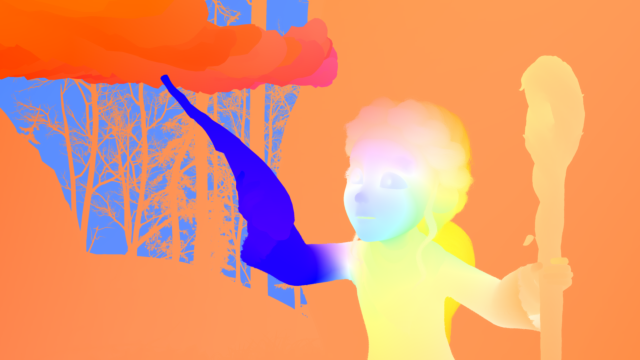}{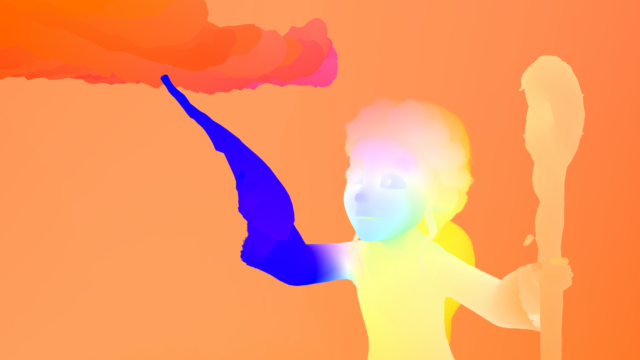}
\\[3pt]
\splitimg{\tabimsize\linewidth}{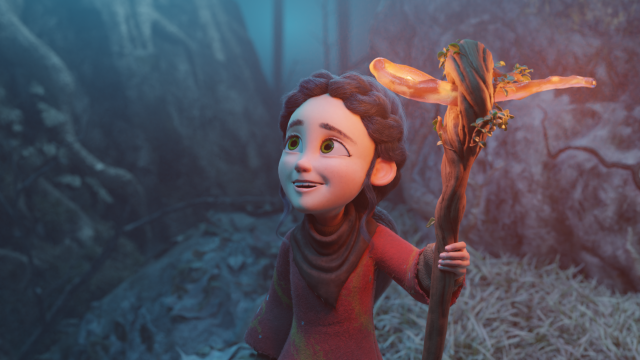}{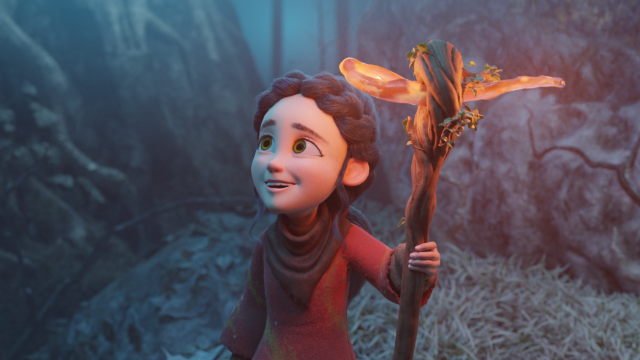}&
\splitimg{\tabimsize\linewidth}{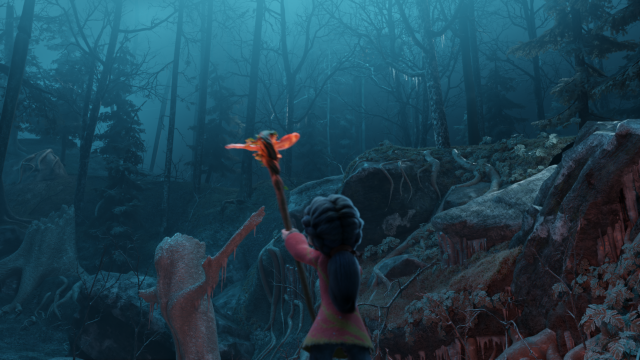}{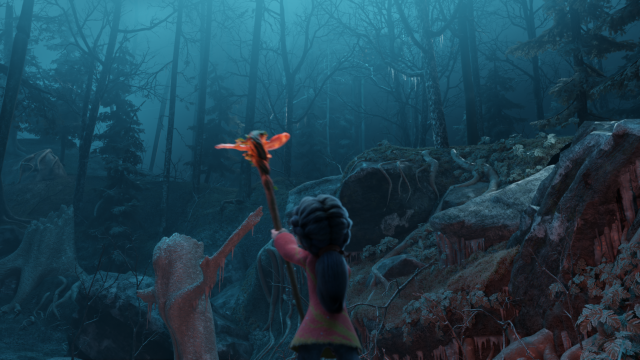}&
\splitimg{\tabimsize\linewidth}{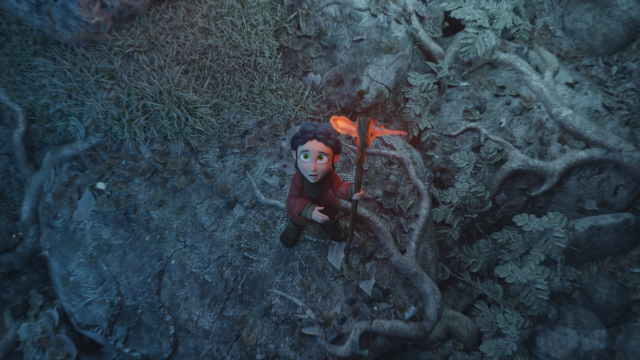}{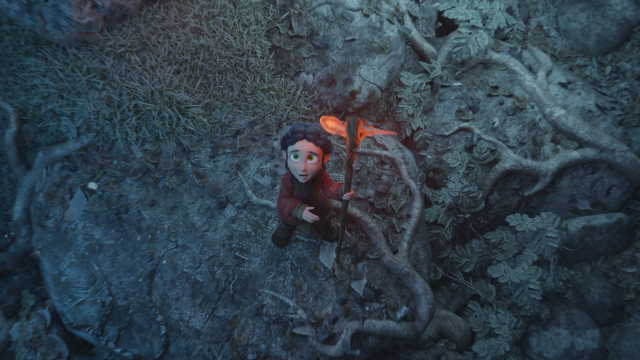}&
\splitimg{\tabimsize\linewidth}{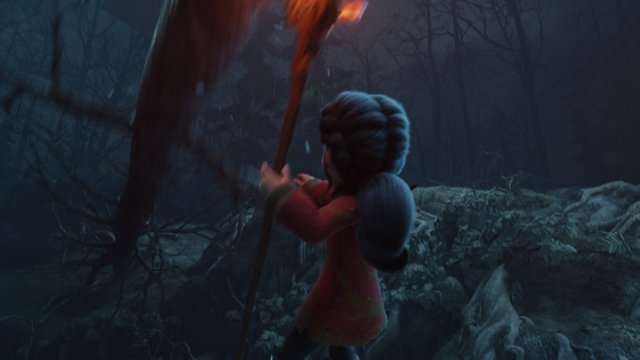}{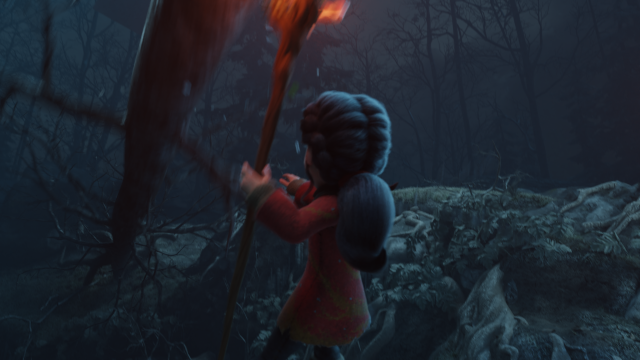}&
\splitimg{\tabimsize\linewidth}{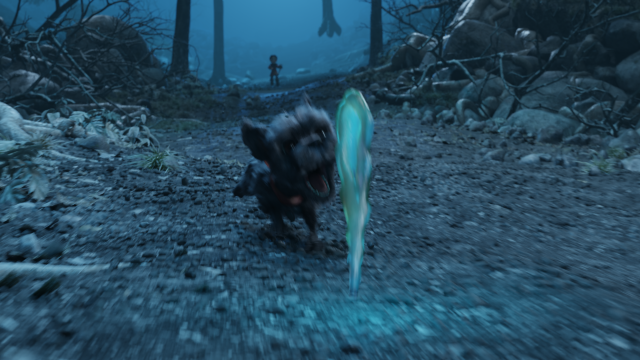}{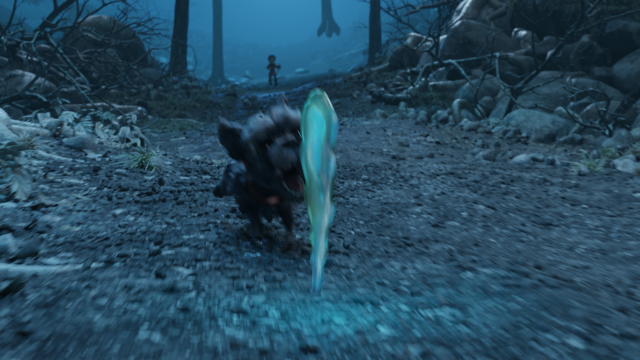}&
\splitimg{\tabimsize\linewidth}{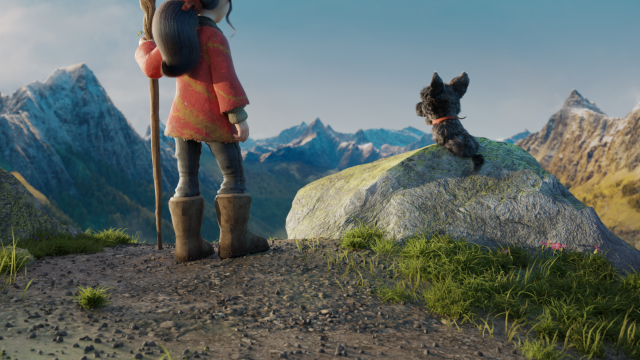}{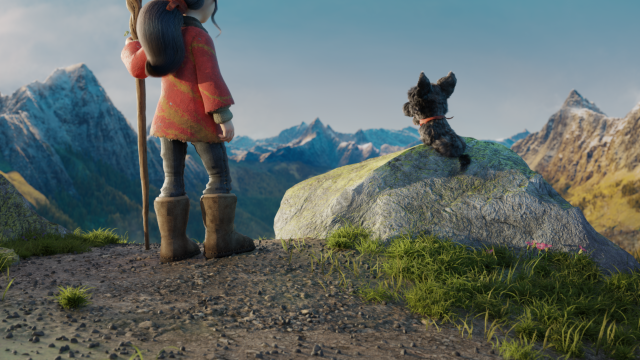}
\\[1pt]
\splitimg{\tabimsize\linewidth}{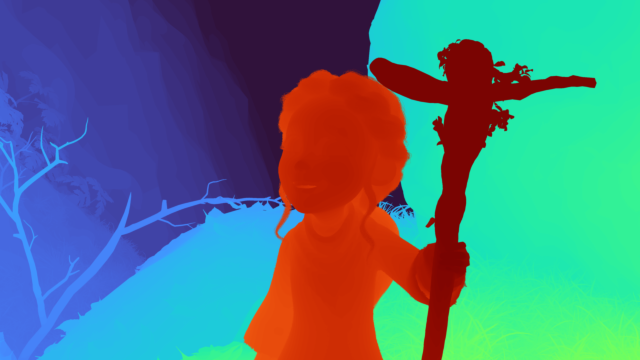}{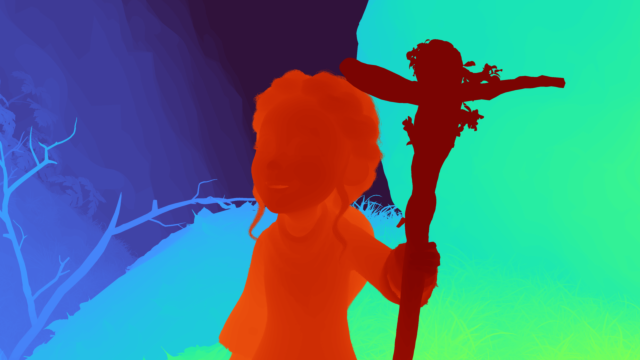}&
\splitimg{\tabimsize\linewidth}{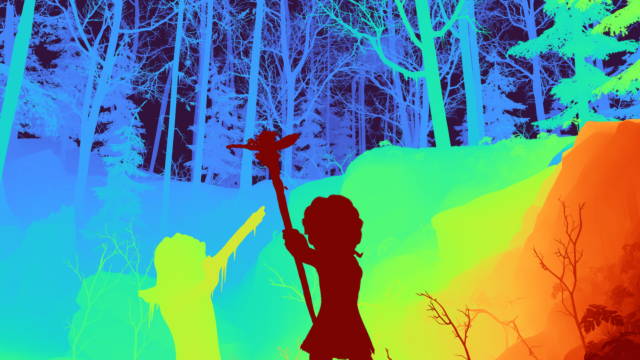}{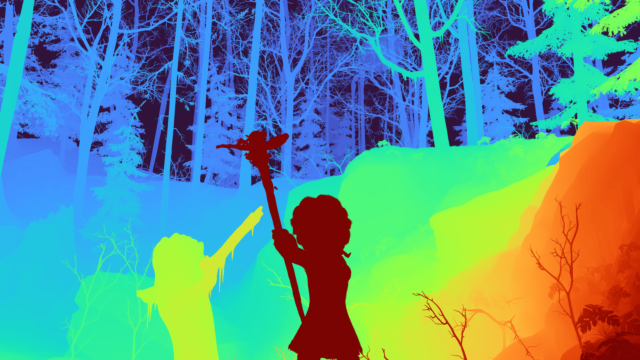}&
\splitimg{\tabimsize\linewidth}{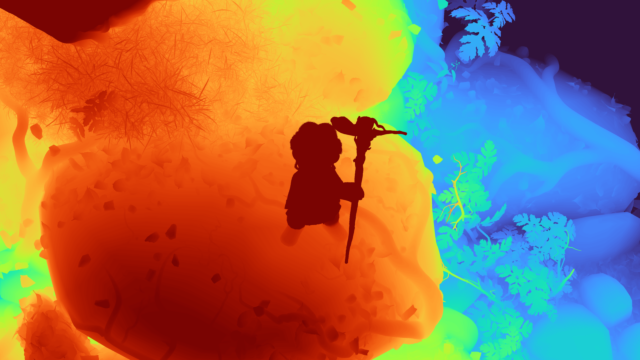}{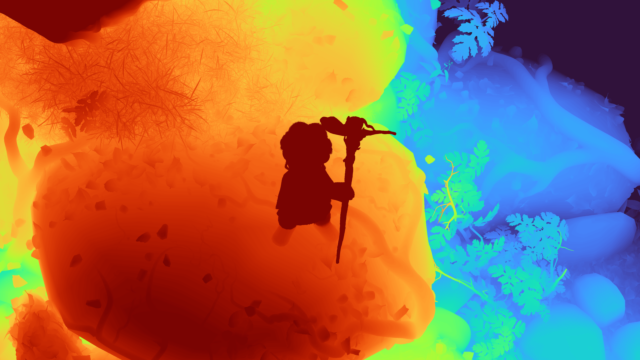}&
\splitimg{\tabimsize\linewidth}{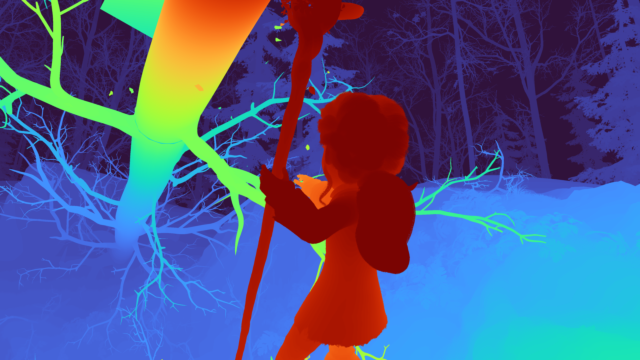}{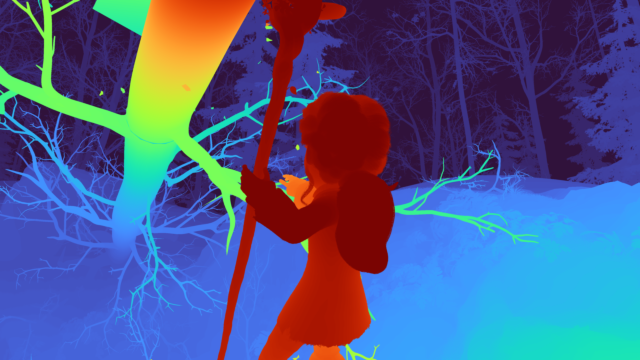}&
\splitimg{\tabimsize\linewidth}{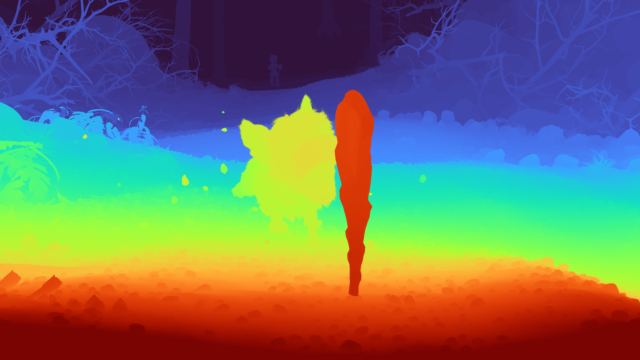}{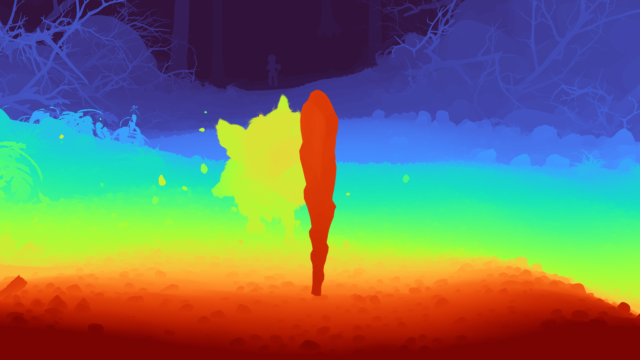}&
\splitimg{\tabimsize\linewidth}{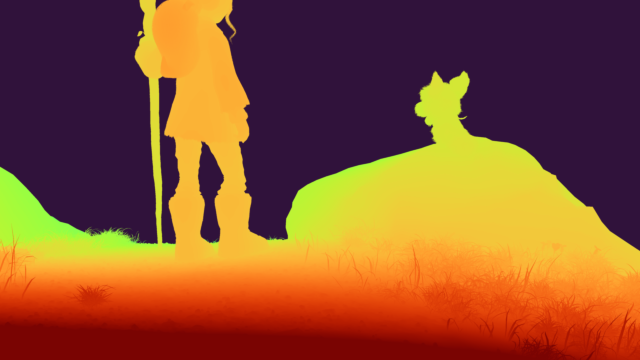}{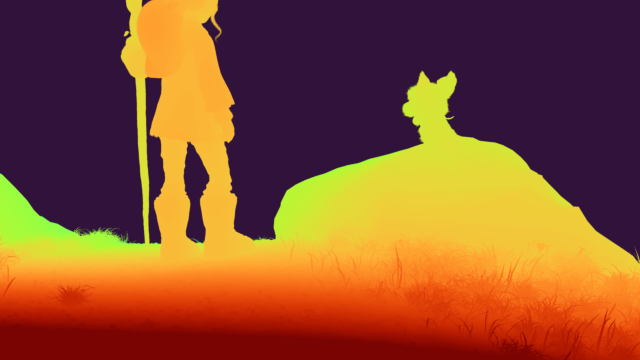}
\\[1pt]
\splitimgfour{\tabimsize\linewidth}{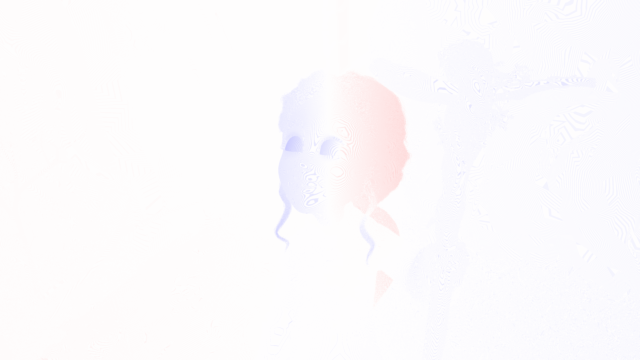}{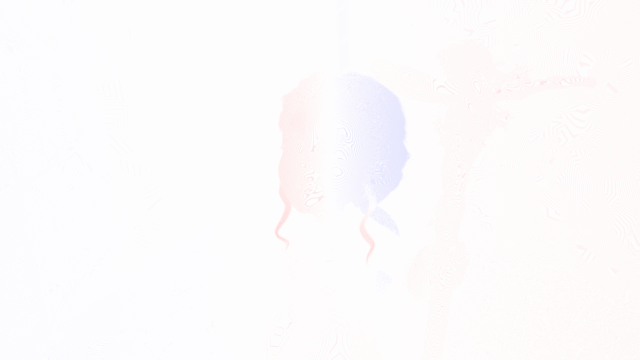}{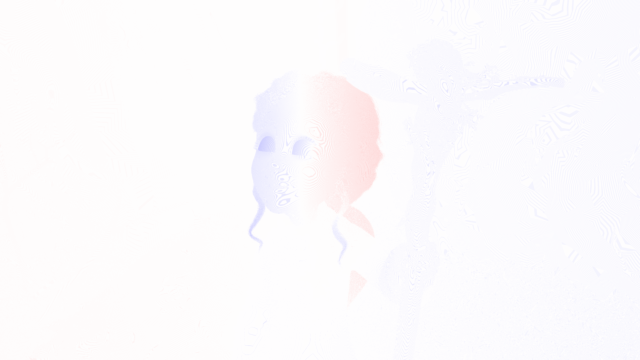}{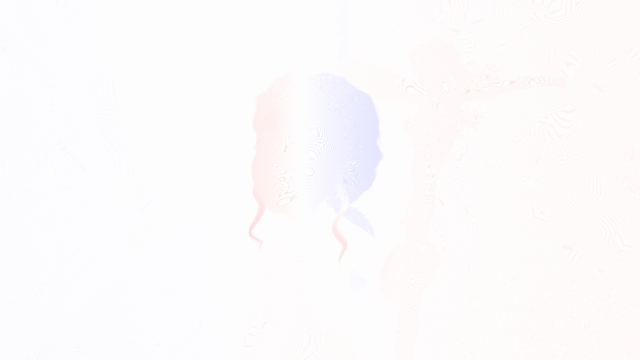}&
\splitimgfour{\tabimsize\linewidth}{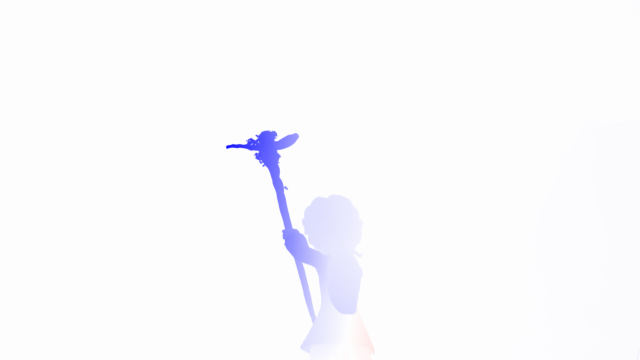}{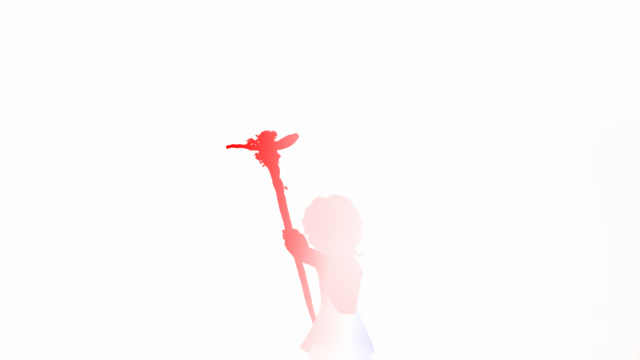}{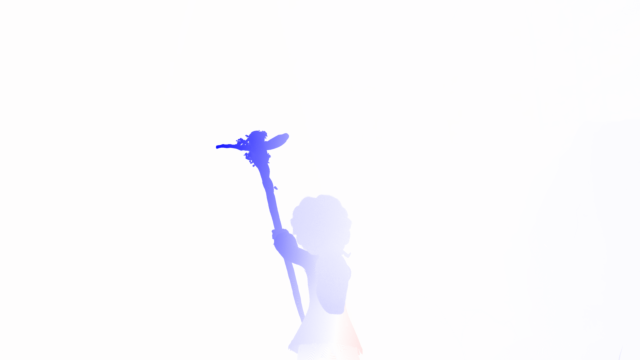}{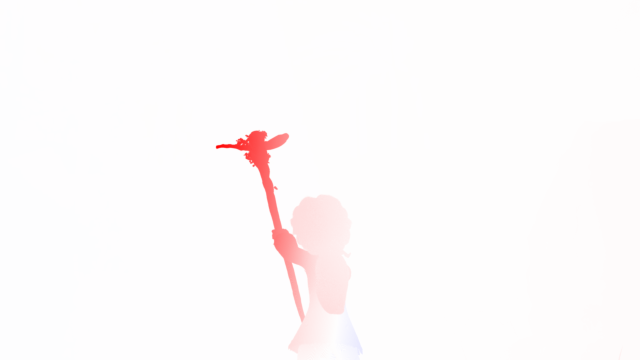}&
\splitimgfour{\tabimsize\linewidth}{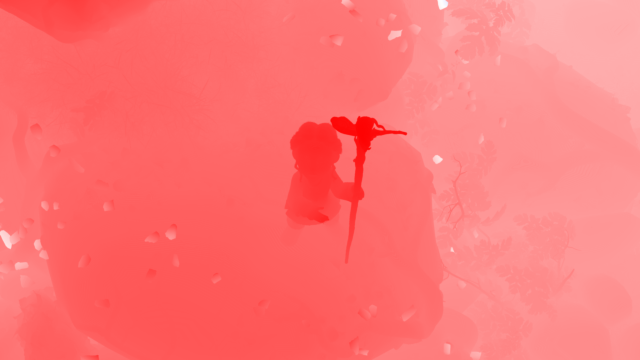}{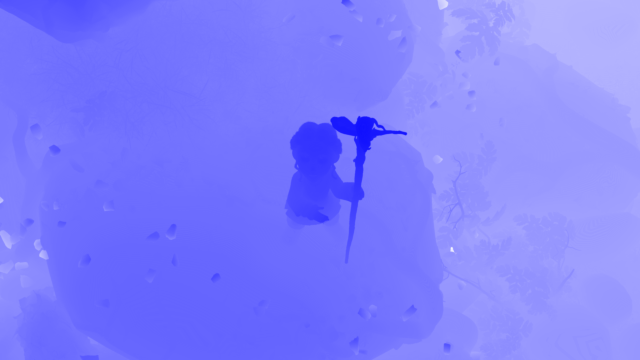}{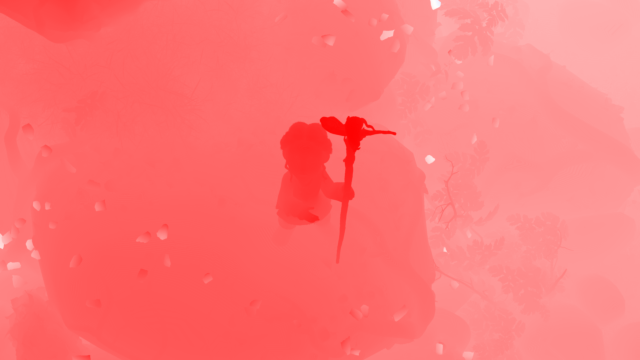}{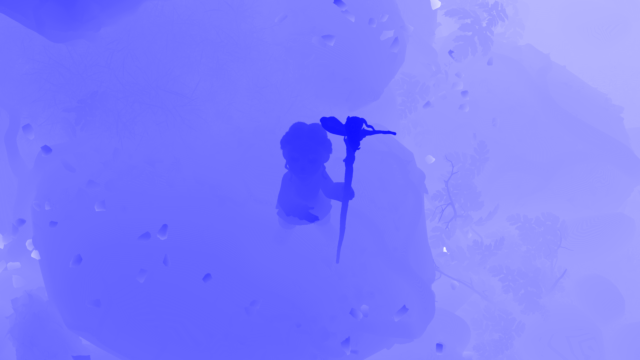}&
\splitimgfour{\tabimsize\linewidth}{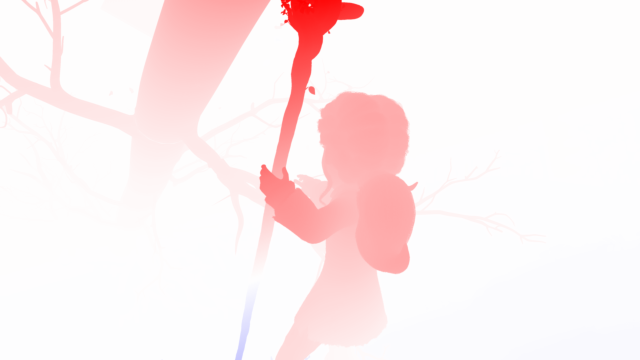}{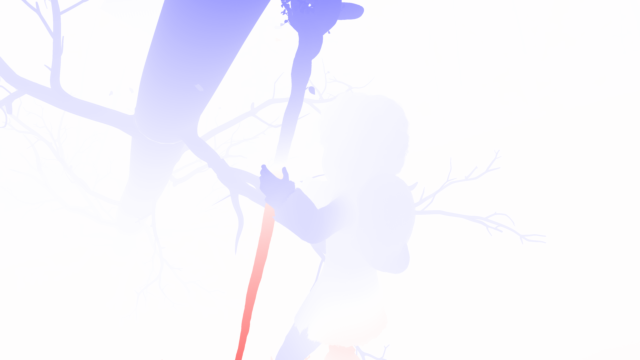}{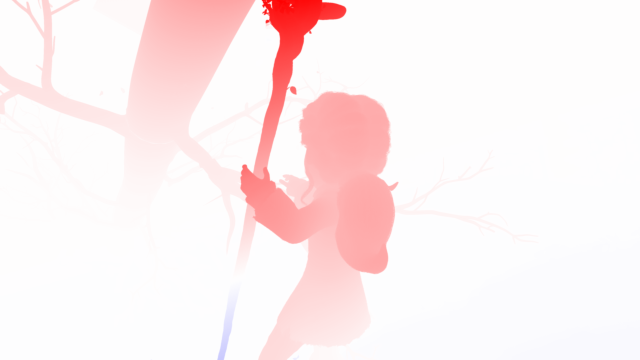}{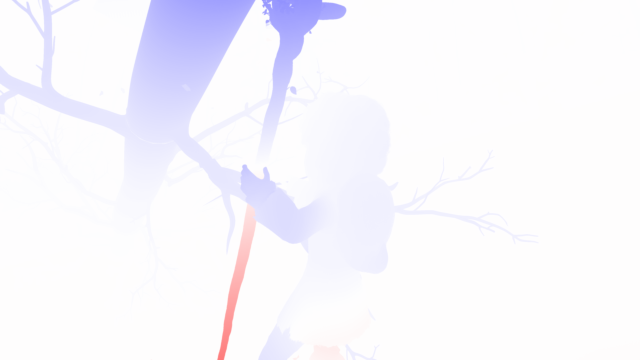}&
\splitimgfour{\tabimsize\linewidth}{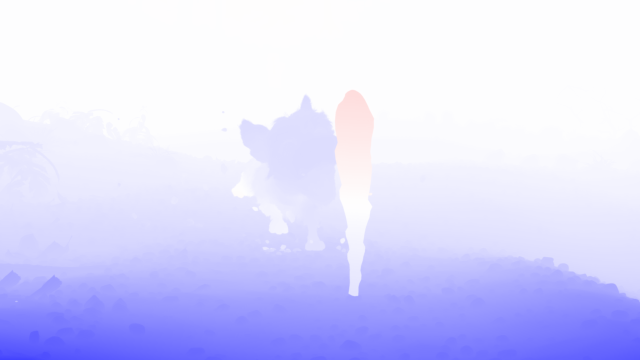}{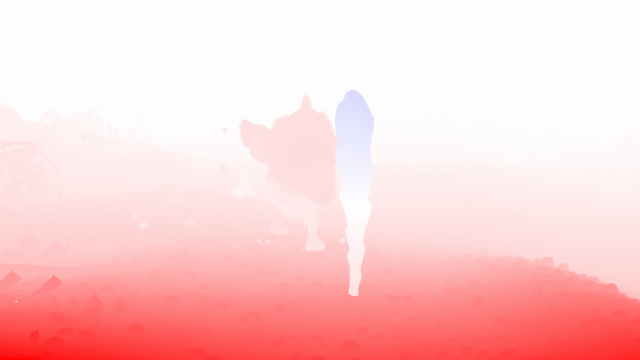}{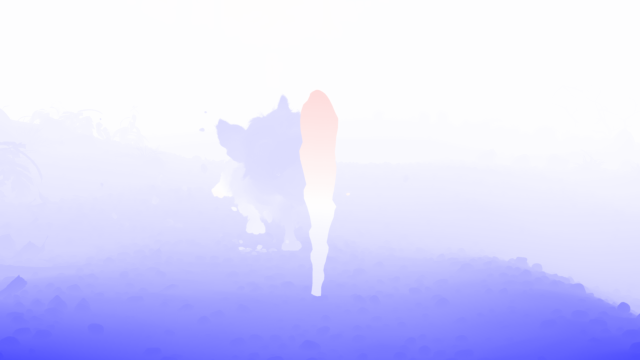}{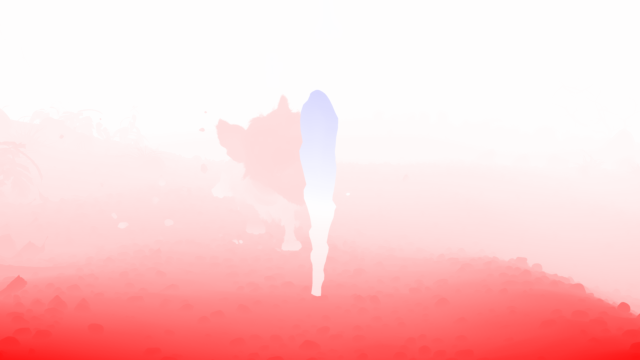}&
\splitimgfour{\tabimsize\linewidth}{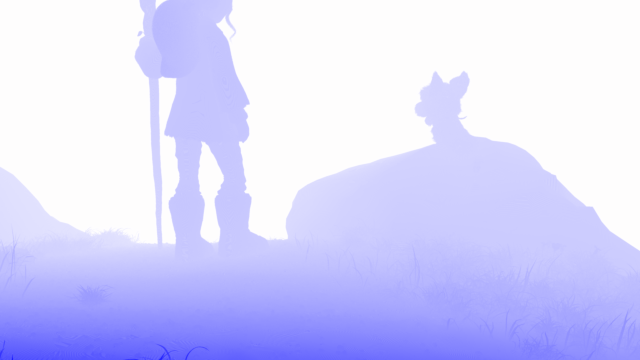}{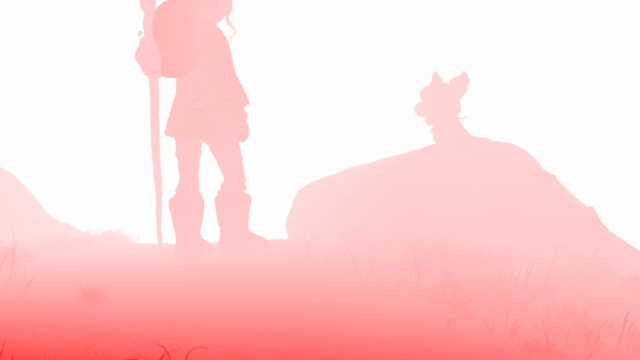}{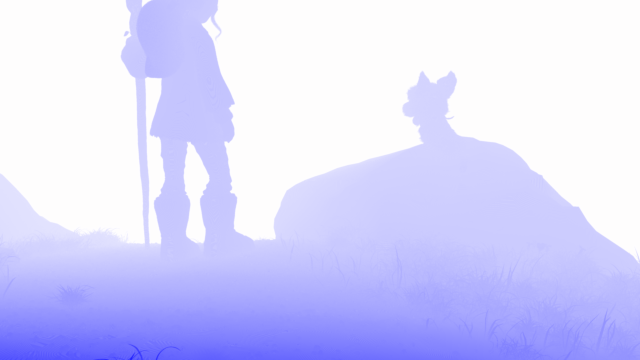}{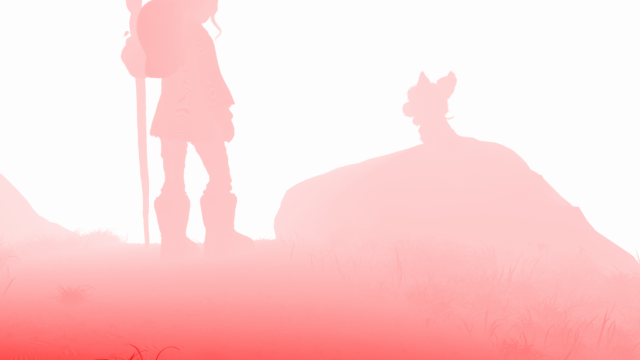}
\\[1pt]
\splitimgfour{\tabimsize\linewidth}{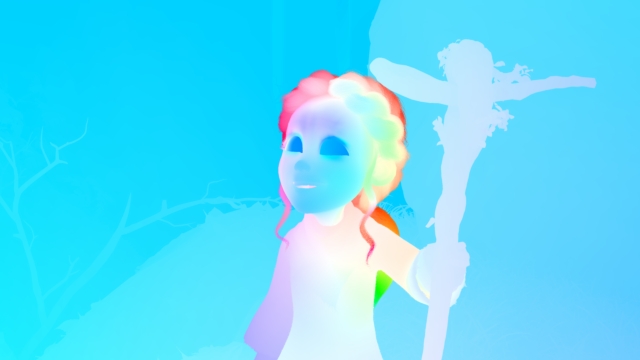}{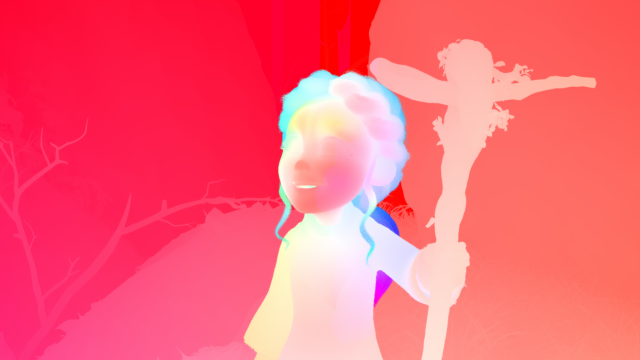}{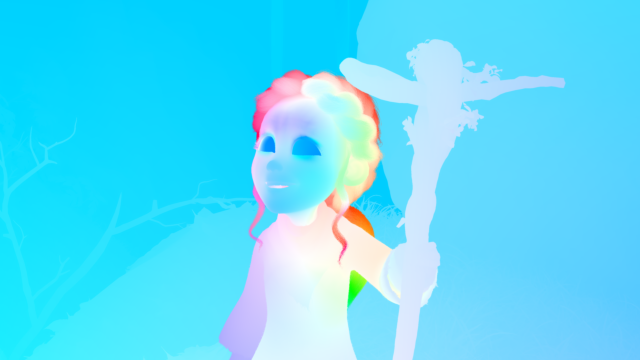}{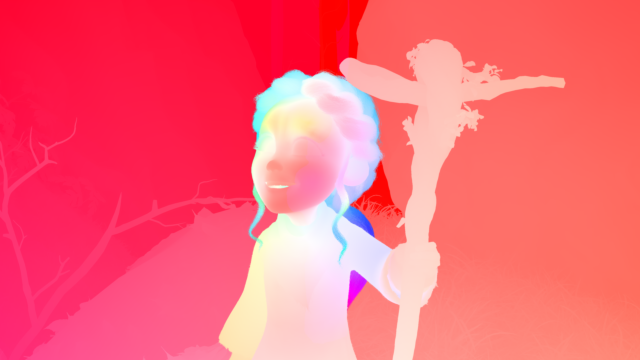}&
\splitimgfour{\tabimsize\linewidth}{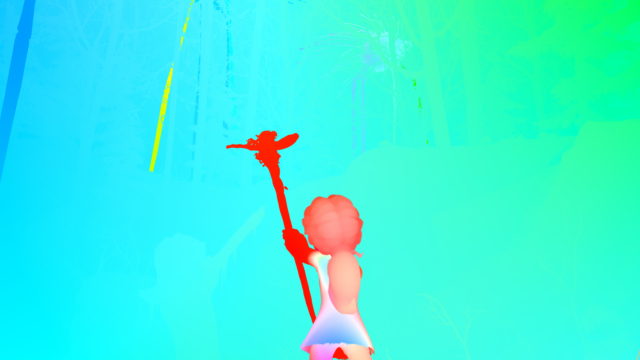}{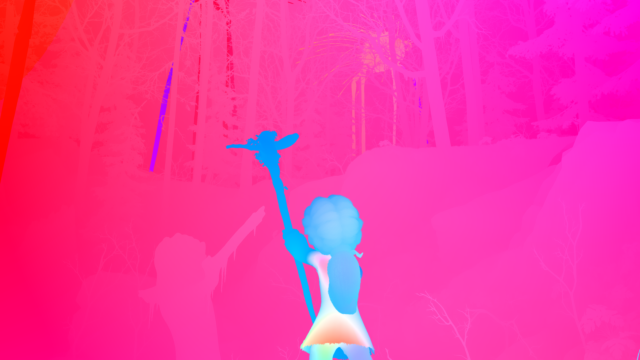}{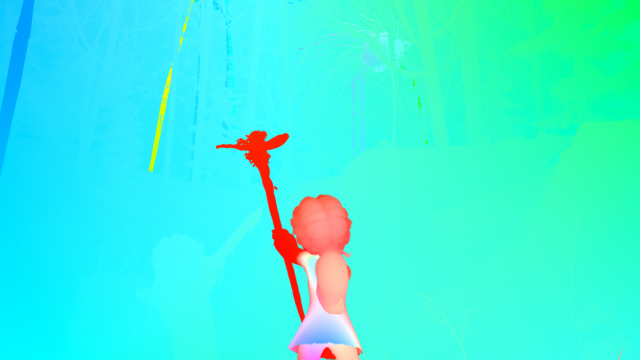}{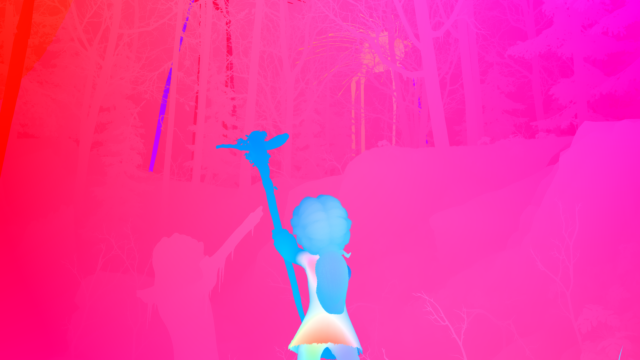}&
\splitimgfour{\tabimsize\linewidth}{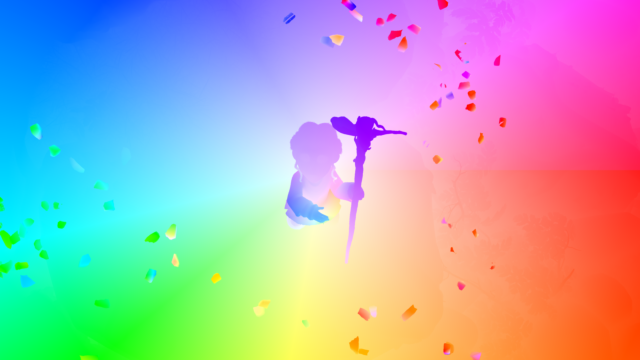}{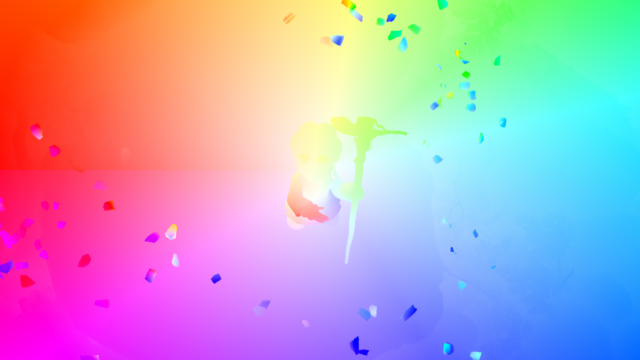}{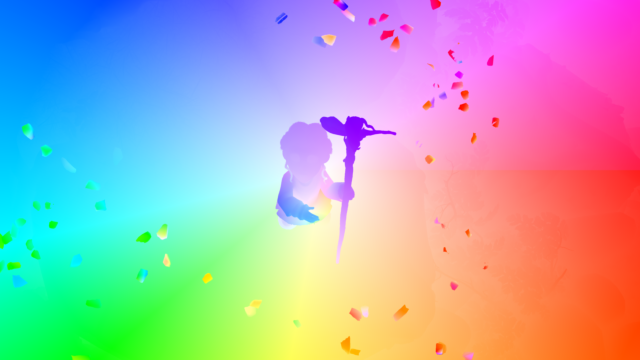}{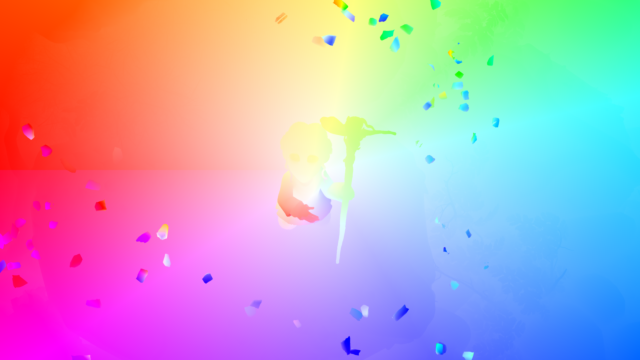}&
\splitimgfour{\tabimsize\linewidth}{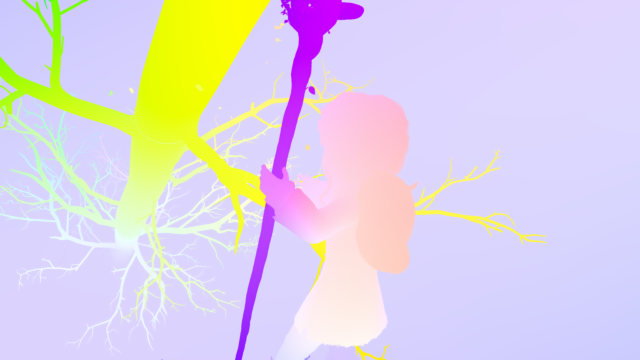}{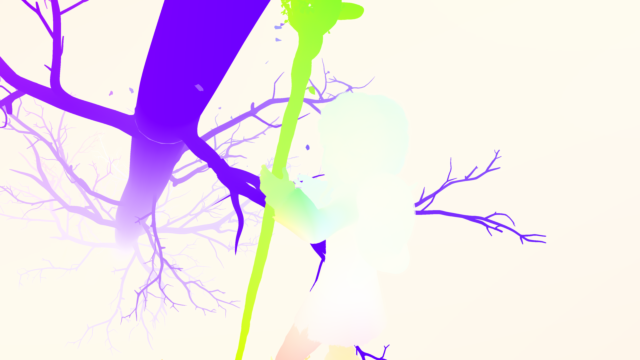}{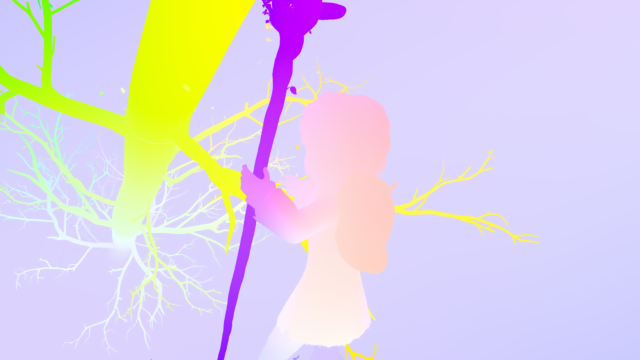}{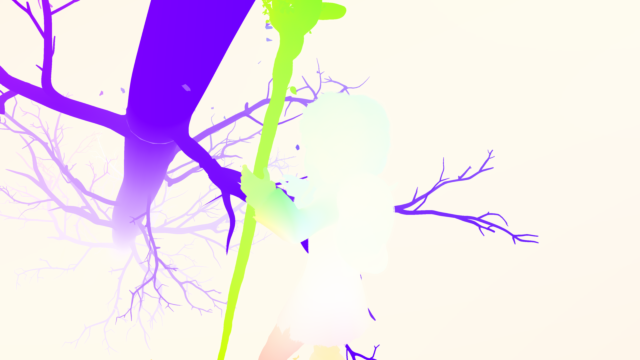}&
\splitimgfour{\tabimsize\linewidth}{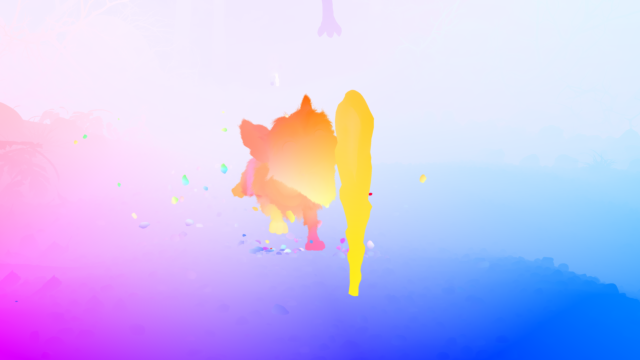}{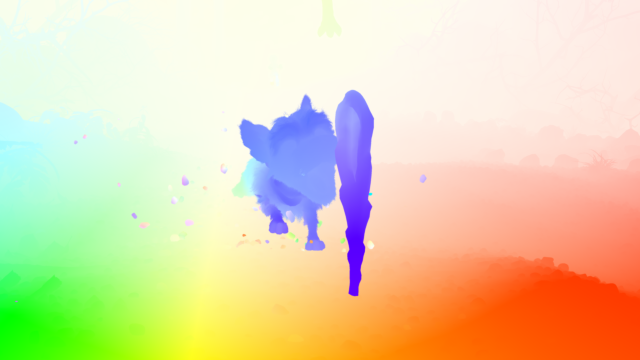}{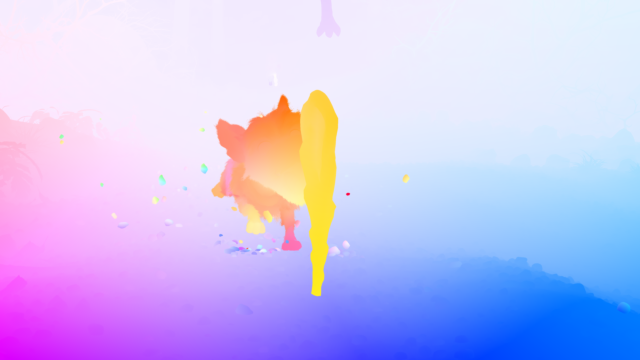}{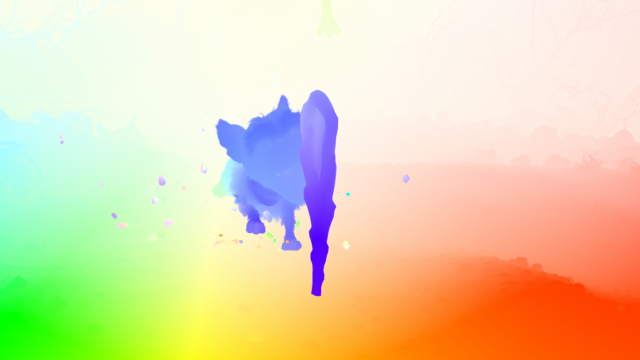}&
\splitimgfour{\tabimsize\linewidth}{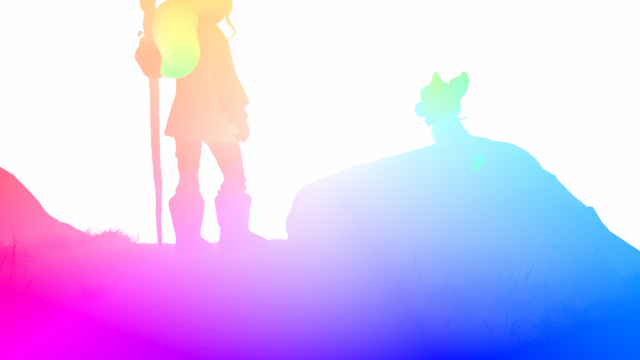}{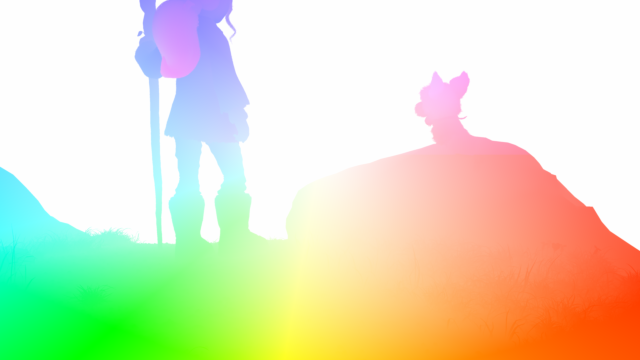}{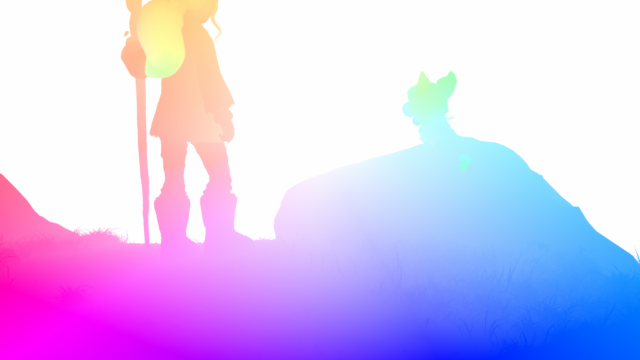}{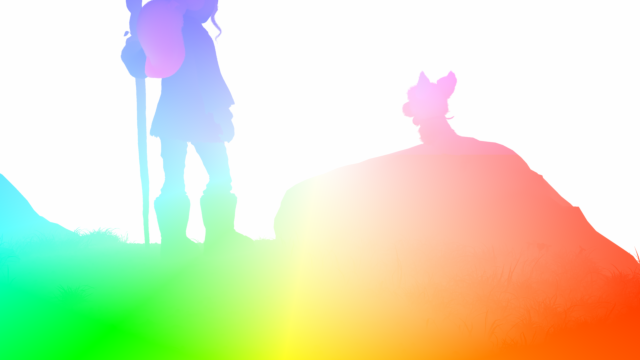}
\end{tabular}
}
\caption{Example sequences from the Spring dataset. \emph{First row}: Left and right images of the stereo camera, \emph{second row}: Corresponding left and right disparity, \emph{third row}: Change in disparity for forward left, backward left, forward right and backward right, \emph{fourth row}: Optical flow visualization for forward left, backward left, forward right and backward right. Please note that we show the disparity change for visualization purposes while the dataset contains the target frame disparity.}
\label{fig:onecol}
\end{figure*}
% \fi

\subsection{Dataset creation}
\label{sec:datasetcreation}
We consider a set of 47 scenes with 6000 frames covering large parts of the original movie.
In each scene, the original monoscopic camera is replaced by a stereoscopic camera with a baseline distance of 6.5 cm.
While keeping the original appearance of the scenes, we had to introduce changes to make them suitable for dataset creation:
Where applicable, we removed dense volumetric clouds, since Blender cannot generate ground truth for them, but kept ambient haze.

We then generated the data in four steps:
First, we rendered the image data in HD resolution with 1920$\times$1080px, including all visual effects.
Second, we generated all data that is required for the ground truth computation, where we disabled motion blur and depth of field and set all objects to solid.
Third, we considered sky regions separately, since Blender is not able to compute ground truth for these regions.
Fourth, we computed additional maps for an improved analysis of results in terms of focused evaluations.
For each step, more details are given below.

\paragraphnew{High-detail structures.}
Up to now, datasets in the literature were not able to represent very thin structures such as hair, grass, or any objects that are smaller than 1px due to the definition of disparities and optical flow as a single representation per pixel.
To mitigate this, \cite{Butler2012_Sintel,Wulff2012_SintelWorkshop} proposed to change the rendering data such that hair is at least 2px wide, which significantly changes the visual appearance.
We propose a novel solution to this problem by generating all ground truth data with twice the spatial resolution, which yields four ground truth values for each image pixel and a total resolution of 3840$\times$2160px.
This way, even fine details like hair or grass are represented in the data, see \cref{fig:details}.
We describe how to use this high-resolution data for evaluation in \cref{sec:evaluation}.

\paragraphnew{Ground truth scene flow from Blender.}
Since it is not possible to directly export 3D motion or scene flow from Blender, we first had to adapt the rendering source code.
We modified it such that we were able to output 3D motion vectors relative to the camera coordinate system.
Then, we extracted forward and backward 3D motion vectors and depth, for both the left and the right camera.
Additionally, we saved the intrinsic and extrinsic camera data as well as the focal distance.
From depth, 3D motion and camera data, it is straightforward to compute ground truth scene flow in the standard parametrization with reference disparity, future/past disparity and optical flow~\cite{Huguet2007}.
We make the code required to generate our dataset with scene flow ground truth from Blender publicly available.

\paragraphnew{Sky areas.}
In several recent datasets, sky areas are not included~\cite{Richter2017_VIPER,Menze2015_KITTI,Geiger2012_KITTI}, yielding a sparse ground truth for many sequences.
Also, the Blender rendering engine is not capable of determining the correct motion vectors for infinitely distant sky points.
Thus, one mitigation strategy is to create a large sphere around the scene~\cite{Mayer2016_FTH,Mayer2020_phdthesis} to obtain motion results for every pixel.
While this provides a reasonable approximation, we opted for an actual \emph{computation} of the ground truth scene flow in sky regions.
For the optical flow, we utilized the relative camera motion to compute the true 2D displacement vectors for pixels of infinite depth, which comes down to considering the rotational part of the relative camera motion only.
For the disparities, the ground truth values are given as 0.
In order to allow for a separate evaluation, we created binary maps determining sky pixels.

\paragraphnew{Additional maps.}
Apart from sky maps, we computed three additional maps to enable a detailed evaluation with our benchmark: high-detail, matching, and rigidity maps; see \cref{fig:maps}.
In the evaluation, we further distinguished between areas of different displacement vector sizes.
In the following, we describe their generation.

A core concept in our dataset is to have the ground truth data available in super-resolution with four ground truth values per pixel.
We use this high-resolution data to compute {\em detail maps} to identify pixels that belong to areas of high details for all scene flow components, \ie optical flow and disparities. %in our dataset 
To this end, we define a point as high-detail if at least one of the four ground truth values deviates from the median of the four ground truth values by more than 1px.

Additionally, we compute {\em matching maps} to differentiate pixels that are matched in the corresponding view from those that are occluded.
This idea has been frequently used~\cite{Geiger2012_KITTI,Menze2015_KITTI,Butler2012_Sintel,Scharstein2002_MiddleburyStereo,Schoeps2017_ETH3D} and is especially important to distinguish regions with matching counterparts from regions where values have to be predicted/extrapolated.
We computed matching maps for all scene flow components through a forward-backward check~\cite{Sundaram2010}.

Further, we calculate {\em rigidity maps} that segment the data into areas where motion is induced only by the camera %motion
and areas where objects move independent of the camera.
Instead of computing rigidity maps in 2D~\cite{Wulff2017}, we determine them in 3D by comparing the ground truth 3D motion vectors to 3D motion vectors that are computed from the static scene and the camera motion.
This 3D strategy prevents errors which are otherwise introduced by comparing projected vectors in 2D.
We consider points to be rigid if the 3D motion vectors differ by at most 1mm.

Finally, we also %compute 
distinguish
areas of different \emph{displacement sizes} for all scene flow components.
Following \cite{Butler2012_Sintel} we select regions of small-size displacements with magnitudes up to 10px (\emph{s0-10}), regions of medium-size displacements with magnitudes of 10-40px (\emph{s10-40}) and regions of even larger displacements exceeding 40px (\emph{s40+}).

\begin{figure*}
    \centering
    \begin{subfigure}[c]{0.239\linewidth}
    \includegraphics[height=3.1cm]{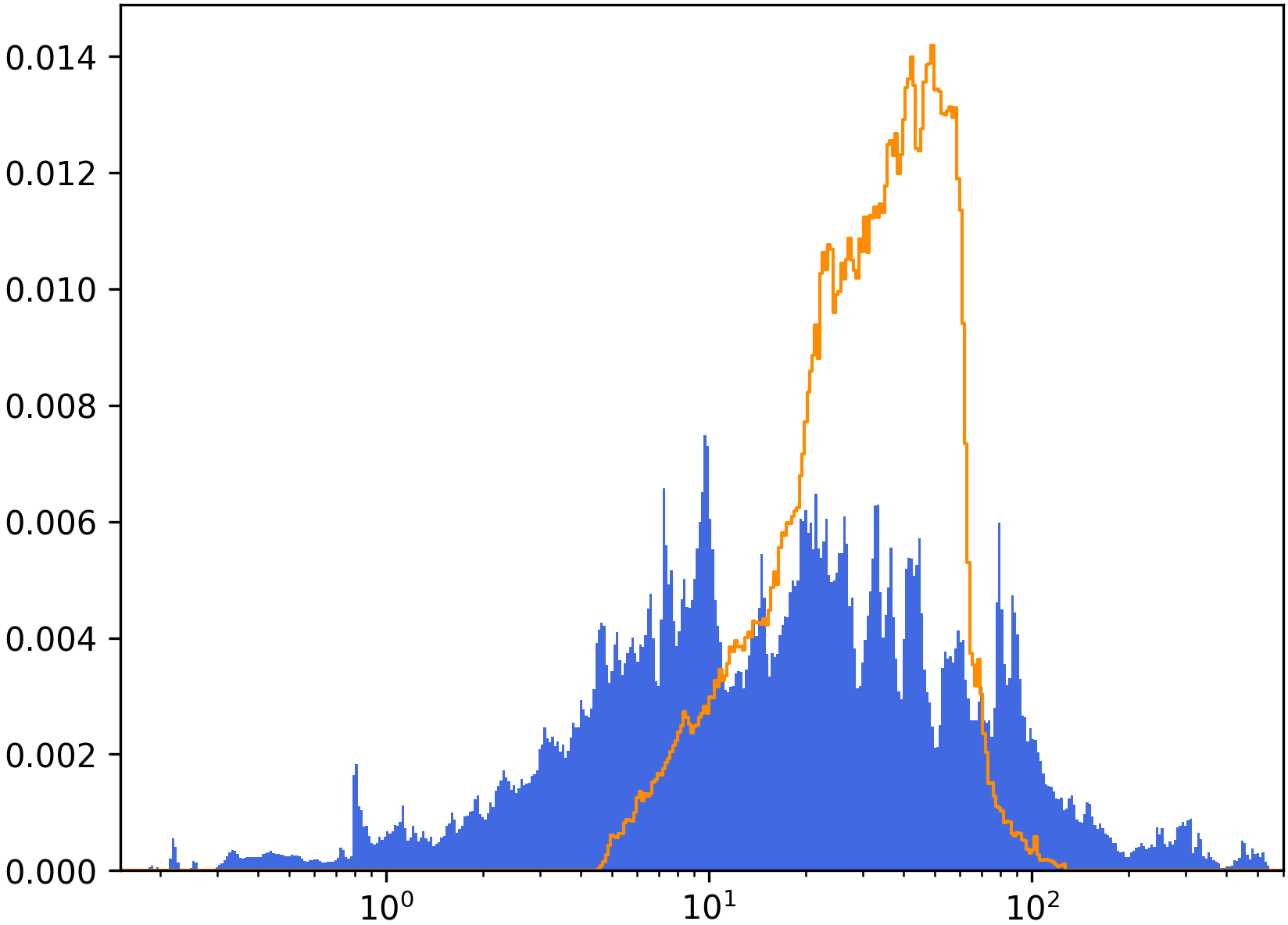}
    \caption{Reference disparity}
    \end{subfigure}
    ~
    \begin{subfigure}[c]{0.239\linewidth}
    \includegraphics[height=3.1cm]{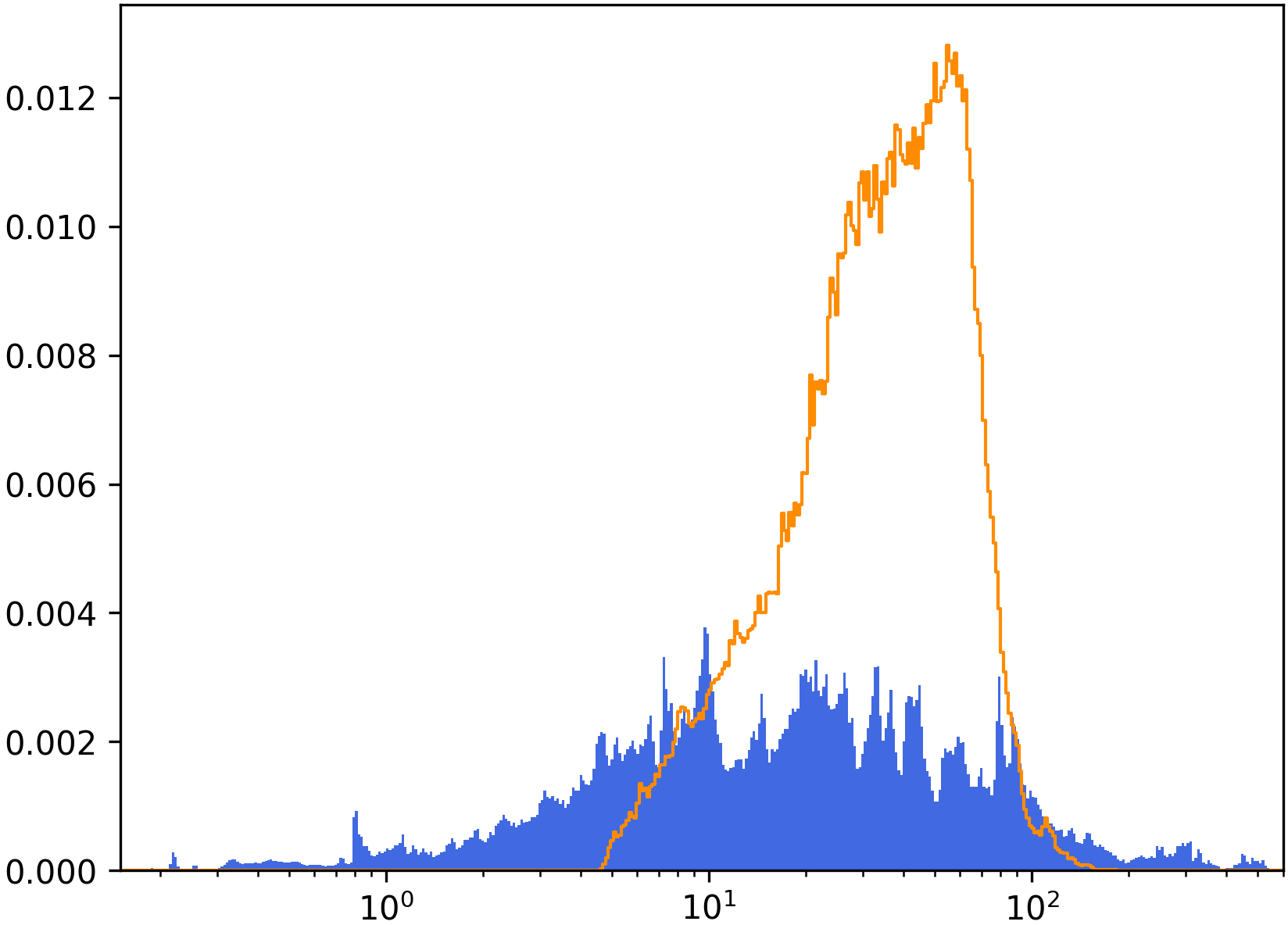}
    \caption{Target disparity}
    \end{subfigure}
    ~
    \begin{subfigure}[c]{0.239\linewidth}
    \includegraphics[height=3.1cm]{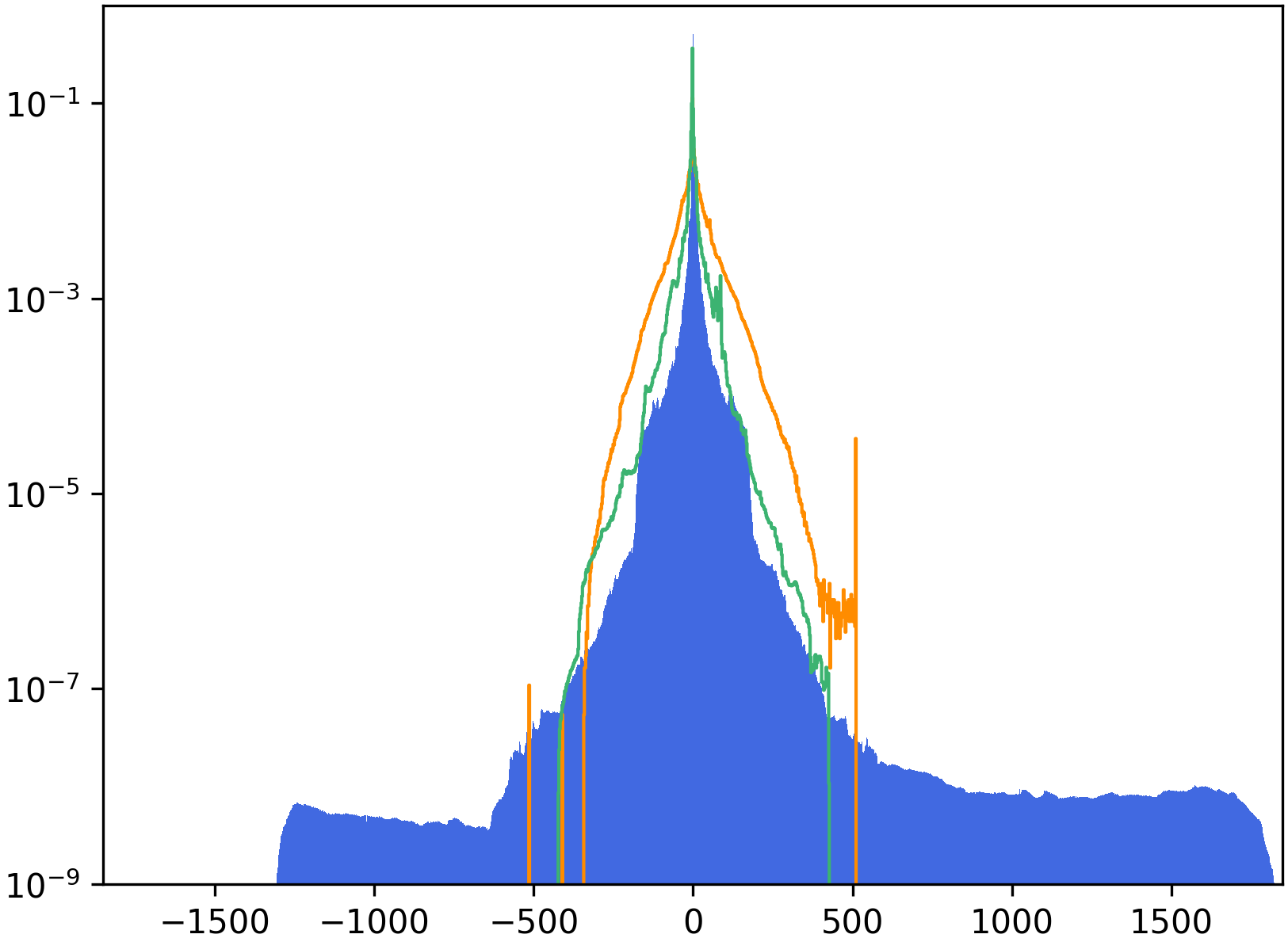}
    \caption{Optical flow $u$ component}
    \end{subfigure}
    ~
    \begin{subfigure}[c]{0.239\linewidth}
    \includegraphics[height=3.1cm]{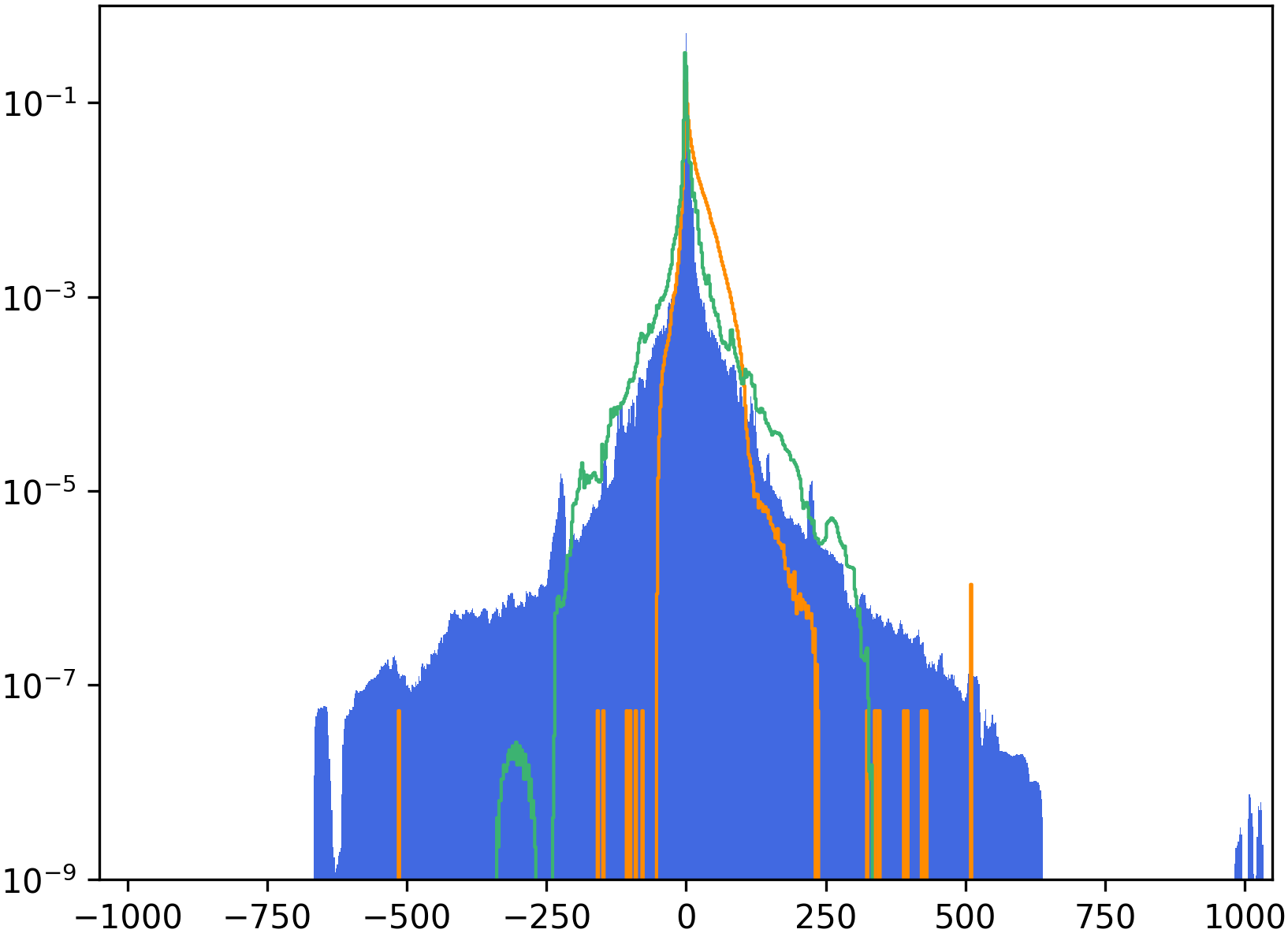}
    \caption{Optical flow $v$ component}
    \end{subfigure}
    \caption{Comparison of ground truth statistics between the datasets Spring (blue) and KITTI 2015 (orange) and Sintel (green).}
    \label{fig:compare}
\end{figure*}

\subsection{Dataset Overview}
Our dataset consists of 47 scenes, which we used to render a total of 6000 stereo frame pairs.
For each left and right frame, we generate disparity ground truth as well as forward and backward optical flow and scene flow ground truth.
In every sequence, we omit the backward flow at the first frame and the forward flow at the last frame.
Thus, considering left/right and forward/backward pairs, our dataset consists of 23812 data samples for scene flow and optical flow and 12000 data samples for stereo estimation.
\Cref{fig:compare} compares the distribution of disparity and optical flow values with the KITTI~15~\cite{Menze2015_KITTI} and MPI Sintel~\cite{Butler2012_Sintel} datasets.
Regarding optical flow, Spring employs a wider range of motion vectors, including very large displacements.
In contrast, for stereo, Spring not only contains larger disparities (very close objects), but also a significant amount of small disparities (far-away objects).

\section{Benchmark}
We split the 47 sequences of our dataset into 37 \emph{train} and 10 \emph{test} sequences, yielding 5000 and 1000 stereo frames, respectively.
We make the full data of the \emph{train} split available, but only publish the images for the \emph{test} split while withholding the ground truth files, which is a standard practise~\cite{Baker2011_MiddleburyFlow,Richter2017_VIPER,Menze2015_KITTI,Geiger2012_KITTI,Butler2012_Sintel,Scharstein2002_MiddleburyStereo}.
In order to allow for a fair comparison, we create a public benchmark website where authors can upload their test split results.
The results are automatically evaluated, with optional display in a public ranking.
Following other benchmarks~\cite{Butler2012_Sintel,Richter2017_VIPER}, we make use of a subsampling strategy to reduce the file size of test split results prior to uploading to the benchmark.
We make the entire code of our benchmark website publicly available.

\begin{table*}[!ht]
\caption{Optical flow results on Spring. %our benchmark. 
We show the 1px outlier rate with sub-rankings for low/high-detail, (un)matched, \mbox{(non-)rigid} and (not) sky regions. Additionally, we show the average endpoint error (EPE), the Fl error~\cite{Menze2015_KITTI} as well as the WAUC metric~\cite{Richter2017_VIPER}.}
\label{tab:opticalflowresults}
\centering
\scalebox{0.85}{
%\begin{adjustbox}{max width=\textwidth}
\setlength\tabcolsep{2pt}
\begin{tabular}{l@{}ccccccccccccccc}
%\begin{tabular}{l@{}G{4pt}{6pt}cG{4pt}{6pt}cG{4pt}{6pt}cG{4pt}{6pt}cG{4pt}{6pt}cG{4pt}{6pt}cG{4pt}{6pt}cG{4pt}{6pt}}
\toprule
& \multicolumn{12}{c}{1px} & \multirow{2}{*}{EPE} & \multirow{2}{*}{Fl} & \multirow{2}{*}{WAUC}
\\
\cmidrule(lr{.75em}){2-13}
Method & total & low-det. & high-det. & matched & unmat. & rigid & non-rig. & not sky & sky & \phantom{0}s0-10 & s10-40 & \;s40+\phantom{0} & & &
\\
\midrule
MS-RAFT+~\cite{Jahedi2022_MSRAFT,Jahedi2022_MSRAFT_RVC} & \phantom{0}5.72 & \phantom{0}5.37 & 61.50 & \phantom{0}5.04 & 33.95 & 3.05 & 25.97 & \phantom{0}4.84 & 19.15 & \phantom{0}2.06 & \phantom{0}5.02 & 38.32 & 0.643 & \phantom{0}2.19 & 92.89
\\
FlowFormer~\cite{Huang2022_Flowformer} & \phantom{0}6.51 & \phantom{0}6.14 & 64.22 & \phantom{0}5.77 & 37.29 & 3.53 & 29.08 & \phantom{0}5.50 & 21.86 & \phantom{0}3.38 & \phantom{0}5.53 & 35.34 & 0.723 & \phantom{0}2.38 & 91.68
\\
FlowNet2~\cite{Ilg2017_Flownet2} & \phantom{0}6.71 & \phantom{0}6.35 & 64.06 & \phantom{0}5.69 & 48.89 & 3.71 & 29.40 & \phantom{0}6.04 & 16.91 & \phantom{0}1.86 & \phantom{0}5.82 & 49.69 & 1.040 & \phantom{0}2.82 & 90.91
\\
RAFT~\cite{Teed2020_RAFT} & \phantom{0}6.79 & \phantom{0}6.43 & 64.09 & \phantom{0}6.00 & 39.48 & 4.11 & 27.09 & \phantom{0}5.25 & 30.18 & \phantom{0}3.13 & \phantom{0}5.30 & 41.40 & 1.476 & \phantom{0}3.20 & 90.92
\\
GMA~\cite{Jiang2021_GMA} & \phantom{0}7.07 & \phantom{0}6.70 & 66.20 & \phantom{0}6.28 & 39.89 & 4.28 & 28.25 & \phantom{0}5.61 & 29.26 & \phantom{0}3.65 & \phantom{0}5.39 & 40.33 & 0.914 & \phantom{0}3.08 & 90.72
\\
GMFlow~\cite{Xu2022_GMFlow}& 10.36 & \phantom{0}9.93 & 76.61 & \phantom{0}9.06 & 63.95 & 6.80 & 37.26 & \phantom{0}8.95 & 31.68 & \phantom{0}5.41 & \phantom{0}9.90 & 52.94 & 0.945 & \phantom{0}2.95 & 82.34
\\
SPyNet\cite{Ranjan2017_SpyNet} & 29.96 & 29.66 & 77.45 & 28.78 & 78.77 & 26.44 & 56.60 & 25.83 & 92.74 & 24.80 & 24.20 & 88.71 & 4.162 & 12.87 & 67.15
\\
PWCNet\cite{Sun2018} & 82.27 & 82.27 & 81.75 & 82.07 & 90.40 & 82.82 & 78.09 & 81.57 & 92.76 & 81.40 & 82.19 & 89.69 & 2.288 & \phantom{0}4.89 & 45.67
\\
\bottomrule
\end{tabular}
}
\end{table*}

\begin{table*}[!ht]
\caption{Stereo results on Spring. %our benchmark. 
We show the 1px outlier rate with sub-rankings for low/high-detail, (un)matched, and (not) sky regions. Additionally, we show the absolute error (Abs), and the D1 error~\cite{Menze2015_KITTI}.}
\label{tab:stereoresults}
\centering
\scalebox{0.85}{
\setlength\tabcolsep{3pt}
\begin{tabular}{lcccccccccccc}
\toprule
& \multicolumn{10}{c}{1px} & \multirow{2}{*}{Abs} & \multirow{2}{*}{D1}
\\
\cmidrule(lr{.75em}){2-11}
Method & total & low-detail & high-detail & matched & unmatched & not sky & sky & s0-10 & s10-40 & s40+
\\
\midrule
ACVNet~\cite{Xu2022_ACVNet} & 14.77 & 14.43 & 35.27 & 12.60 & 57.89 & 11.16 & 69.62 & 18.39 & 11.35 & 18.15 & 1.52 & \phantom{0}5.35
\\
RAFT-Stereo~\cite{Lipson2021_RAFTStereo} & 15.27 & 14.98 & 32.77 & 13.39 & 52.58 & \phantom{0}9.92 & 96.57 & 22.59 & 10.02 & 17.09 & 3.02 & \phantom{0}8.63
\\
LEAStereo\cite{Cheng2020_LEAStereo} & 19.89 & 19.55 & 40.40 & 17.61 & 65.09 & 16.73 & 67.81 & 19.08 & 13.86 & 39.41 & 3.88 & \phantom{0}9.19
\\
GANet~\cite{Zhang2019_GANet} & 23.22 & 22.91 & 42.06 & 20.98 & 67.88 & 18.42 & 96.27 & 24.29 & 16.43 & 41.50 & 4.59 & 10.39
\\
\bottomrule
\end{tabular}
}
\end{table*}

\begin{table*}[!ht]
\caption{Scene flow results on Spring. %our benchmark. 
We show the 1px outlier rate with sub-rankings for low/high-detail, (un)matched, (non-)rigid and (not) sky regions. Additionally, we show the SF error~\cite{Menze2015_KITTI} as well as individual 1px outlier rates for reference disparity (1px$^\text{D1}$), target disparity (1px$^\text{D2}$) and optical flow (1px$^\text{Fl}$).}
\label{tab:sceneflowresults}
\centering
%\begin{adjustbox}{max width=\textwidth}
\scalebox{0.85}{
\setlength\tabcolsep{2pt}
\begin{tabular}{l@{}cccccccccccccccc}
\toprule
& \multicolumn{12}{c}{1px} & \multirow{2}{*}{SF} & \multirow{2}{*}{1px$^\text{D1}$} & \multirow{2}{*}{1px$^\text{D2}$} & \multirow{2}{*}{1px$^\text{Fl}$}
\\
\cmidrule(lr{.75em}){2-13}
Method & total & low-det. & high-det. & matched & unmat. & rigid & non-rig. & not sky & sky & s0-10 & s10-40 & s40+
\\
\midrule
M-FUSE (F)~\cite{Mehl2023_MFUSE} & 34.90 & 34.30 & 64.32 & 32.03 & 71.94 & 29.81 & 73.38 & 31.36 & 88.71 & 29.89 & 23.91 & 69.15 & 16.10 & 19.89 & 24.26 & 20.37
\\
RAFT-3D (K)~\cite{Teed2021_RAFT3D} & 37.26 & 36.80 & 60.23 & 34.34 & 75.02 & 32.87 & 70.52 & 33.23 & 98.53 & 43.80 & 24.55 & 63.91 & 17.35 & 32.31 & 32.95 & 13.96
\\
CamLiFlow (F)~\cite{Liu2022} & 50.08 & 49.64 & 71.88 & 47.80 & 79.64 & 46.75 & 75.27 & 46.85 & 99.25 & 31.12 & 42.70 & 89.55 & 34.15 & 23.22 & 44.10 & 24.01
\\
M-FUSE (K)~\cite{Mehl2023_MFUSE} & 62.49 & 62.29 & 72.31 & 60.57 & 87.25 & 60.39 & 78.42 & 60.03 & 99.85 & 49.20 & 75.96 & 25.23 & 25.23 & 52.23 & 57.03 & 20.98
\\
RAFT-3D (F)~\cite{Teed2021_RAFT3D} & 78.82 & 78.75 & 82.20 & 78.33 & 85.19 & 79.62 & 72.80 & 77.57 & 97.84 & 84.33 & 81.68 & 65.48 & 66.88 & 23.22 & 73.43 & 48.07
\\
CamLiFlow (K)~\cite{Liu2022}~~ & 85.31 & 85.18 & 91.67 & 84.46 & 96.25 & 84.18 & 93.85 & 84.35 & 99.96 & 65.16 & 87.85 & 99.86 & 70.87 & 32.31 & 76.32 & 69.68
\\
\bottomrule
\end{tabular}
%\end{adjustbox}
}
\end{table*}

\begin{table}
\caption{Influence of the subsampling on the evaluation. We compare optical flow results evaluated using the subsampling of the benchmark with the same results evaluated on the full \emph{test} split.}
% for several widely used error measures.}
\label{tab:subsampling}
\centering
%\begin{adjustbox}{max width=\linewidth}
\scalebox{0.85}{
\setlength\tabcolsep{2.5pt}
\begin{tabular}{l@{}ccccccc}
\toprule
& \multicolumn{3}{c}{subsampling results} && \multicolumn{3}{c}{full \emph{test} split results}
\\
\cmidrule(lr{.25em}){2-4}
\cmidrule(lr{.15em}){6-8}
& 1px & EPE & Fl & & 1px & EPE & Fl
\\
\midrule
MS-RAFT+~\cite{Jahedi2022_MSRAFT,Jahedi2022_MSRAFT_RVC}~~ & \phantom{0}5.72 & 0.643 & \phantom{0}2.19 && \phantom{0}4.99 & 0.620 & \phantom{0}1.82
\\
FlowFormer~\cite{Huang2022_Flowformer} & \phantom{0}6.51 & 0.723 & \phantom{0}2.38 && \phantom{0}6.12 & 0.719 & \phantom{0}2.18
\\
FlowNet2~\cite{Ilg2017_Flownet2} & \phantom{0}6.71 & 1.040 & \phantom{0}2.82 && \phantom{0}5.91 & 0.968 & \phantom{0}2.30
\\
RAFT~\cite{Teed2020_RAFT} & \phantom{0}6.79 & 1.476 & \phantom{0}3.20 && \phantom{0}6.05 & 1.265 & \phantom{0}2.72
\\
GMA~\cite{Jiang2021_GMA} & \phantom{0}7.07 & 0.914 & \phantom{0}3.08 && \phantom{0}6.30 & 0.918 & \phantom{0}2.69
\\
GMFlow~\cite{Xu2022_GMFlow}& 10.36 & 0.945 & \phantom{0}2.95 && \phantom{0}9.38 & 0.928 & \phantom{0}2.49
\\
SPyNet\cite{Ranjan2017_SpyNet} & 29.96 & 4.162 & 12.87 && 28.72 & 4.036 & 11.95
\\
PWCNet~\cite{Sun2018} & 82.27 & 2.288 & \phantom{0}4.89 && 82.17 & 2.295 & \phantom{0}4.47
\\
\bottomrule
\end{tabular}
%\end{adjustbox}
}
\end{table}

\subsection{Evaluation metrics}
\label{sec:evaluation}
As described in \cref{sec:dataset}, we generate the ground truth data in double resolution, resulting in four ground truth values per pixel.
Out of these, the evaluation always selects the ground truth value closest to the estimated value for calculating the errors.
%In the evaluation, we use this data by always selecting the ground truth value that is closest to the estimated value.
In the case of of thin hair structures against a background, this strategy yields a low error for methods that estimate the hair or the background value, while assigning a larger error when a mixture of both values is predicted.

In the literature, there is a multitude of error measures for scene flow, optical flow and stereo methods.
For scene flow, the most established ones are the outlier rates by KITTI. 
They define pixels as outliers if they deviate by more than 3px and 5\% from the ground truth, which is motivated from the limited precision of their data~\cite{Menze2015_KITTI}.
%We adapt their evaluation to our high-accuracy data and consider
Considering our high-accuracy data, we adapt the evaluation to
%We simplify their evaluation approach to 
the \emph{1px outlier rate}.
For reference disparity, target disparity and optical flow, the 1px outlier rate defines the percentage of pixels that deviate more than 1px from the ground truth.
Following \cite{Menze2015_KITTI}, we also employ a \emph{union} 1px error as the main scene flow measure that defines the percentage of pixels where \emph{any} of the estimated reference disparity, target disparity and optical flow values deviates more than 1px from the ground truth.
Since for scene flow, optical flow and stereo no single error measure is fully established in the community, we also provide multiple error metrics as additional reference.
For stereo, we also show the absolute error and the KITTI-D1 and D2 error~\cite{Menze2015_KITTI}; for optical flow, we show the end point error, the KITTI-Fl error and the WAUC error~\cite{Richter2017_VIPER}; for scene flow, we show all previously mentioned errors as well as the KITTI-SF error~\cite{Menze2015_KITTI}.
To enable an in-depth analysis, we make use of the maps described in \cref{sec:datasetcreation} and additionally report sub-errors for different parts of the scene.

\subsection{Rules and cheating prevention}
Our cheating prevention strategy is threefold.
First, we require authors to register on our website giving their affiliation and a brief justification why they need access to the benchmark.
After verification of the account, which shall prevent mass-registration, authors are allowed to submit results under upload limits \cite{Butler2012_Sintel,Menze2015_KITTI}: At most once per hour and three times per 30 days. This way, we prevent overfitting on the test data.
Second, we added several sequences with manual adjustment of the camera path~\cite{Butler2012_Sintel} in our data generation process to prevent users from utilizing the publicly available movie assets.
Third, we make our full benchmark code publicly available, but hide the exact subsampling used for generating the submission files by only providing compiled subsampling executables.

\subsection{Initial results}
For evaluation on our benchmark, we consider %a set of 
15 state-of-the-art methods
(8 optical flow, 4 stereo, 3 scene flow)
whose code is publicly available. % and evaluate them on our benchmark.
The results are given in \cref{tab:sceneflowresults,tab:opticalflowresults,tab:stereoresults}.
%For the evaluation, 
We selected author-provided checkpoints trained on Sintel for optical flow, on FlyingThings3D for stereo, and on both FlyingThings3D and KITTI for scene flow.
We argue that, at the initial stage, an evaluation of methods that are not finetuned on Spring results in a fairer comparison than retraining methods of other authors on the \emph{train} split of our dataset. Moreover, using non-finetuned methods also gives interesting insights %regarding 
into the generalization performance of existing methods to novel benchmarks. At the same time, it is clear that these results and rankings can only serve as starting point. Hence, we 
%would like to 
encourage authors to submit finetuned versions of their methods to our benchmark.
In the following, we give an initial discussion of our %analysis 
results, which is extended in the supp.\ %lementary 
material.

\paragraphnew{Optical flow.} 
For optical flow, we can observe that although recent methods perform generally well, errors in areas with high details are still very large -- although our novel evaluation method is quite permissive in those areas.
Furthermore, we find that the errors in unmatched, non-rigid, sky and large-displacement areas are also large with the error in the sky being actually the smallest.
Among the different methods,
MS-RAFT+
\cite{Jahedi2022_MSRAFT_RVC}, which is specifically tailored %tuned 
for high resolutions and single-checkpoint cross-benchmark generalization \cite{rvc}, ranks first.
Moreover, the classical Flow\-Net2 \cite{Ilg2017_Flownet2} performs surprisingly well which can be attributed to its dedicated module for small displacement estimation
(\cf the \emph{s0-10} metric).
In contrast, the well-established PWCNet~\cite{Sun2018} ranks last. 
While it provides reasonable results in terms of the EPE and Fl error, 
its accuracy seems to be limited by its strategy to operate on 1/4 of the input resolution and subsequently upsample the flow using simple bilinear interpolation.
Overall, these observations demonstrate that
for Spring, the capability of the underlying architecture 
to handle high-detail high-resolution input is much more important than for other benchmarks. % \cite{Butler2012_Sintel,Menze2015_KITTI}. 
This in turn outlines the value of Spring when it comes to further pushing the limits of current optical flow methods.

\paragraphnew{Stereo.} 
For stereo estimation,
we can see that results are generally worse compared to optical flow.
While surprising at first sight, we attribute this to mainly two reasons.
First, most stereo methods consider very clean data, \ie data without camera defocus and/or motion blur, which stands in contrast to our dataset.
Second, stereo disparity is often defined to be strictly positive~\cite{Geiger2012_KITTI,Menze2015_KITTI,Mayer2016_FTH}, thus in general stereo methods are not prepared for regions with zero disparity/infinite depth as for the sky in our dataset -- which can be seen in the corresponding sub-metric, as well as the \emph{s0-10} metric.
However, we argue that disparity methods, and subsequently scene flow methods, should be able to predict true dense fields, including sky regions.
For reference, we provide a full evaluation solely on non-sky pixels for all methods including optical flow and scene flow in the %supplementary 
supp.\ material.
As a final note, previous stereo benchmarks often report results by default in non-occluded (matched) regions only~\cite{Schoeps2017_ETH3D,Geiger2012_KITTI,Scharstein2002_MiddleburyStereo,Scharstein2014_MiddleburyStereo}, which drives the development of methods that perform especially well in these areas.
All in all, these results clearly demonstrate the advantage of the Spring benchmark for the field of stereo estimation.

\paragraphnew{Scene flow.} 
For scene flow estimation, we evaluated two trained models for each method, corresponding to a pre-training
on FlyingThings3D (F) and a subsequent finetuning on KITTI 2015 (K).
Since the considered scene flow methods strongly rely on a preceding disparity estimation, the same observations hold as in case of the stereo results.
Furthermore, the inconsistent results of all three methods with their two training schedules show that solely focusing on the KITTI 2015 benchmark produces methods that are prone to overfitting and lack a good generalization performance.
Hence, the Spring benchmark is also highly beneficial to advance research in the field of scene flow.

\subsection{Influence of subsampling}
\label{sec:subsampling}
As previously outlined, our benchmark uses a subsampling strategy that evaluates on a reduced set of ground truth pixels from the full \emph{test} set.
In a final experiment, we investigate the influence of this strategy by comparing the subsampling results shown in \cref{tab:opticalflowresults} to results computed on the full \emph{test} split.
This is the first time %~\cite{Butler2012_Sintel,Richter2017_VIPER} 
in the dense matching literature that the influence of the subsampling evaluation is made transparent.
\Cref{tab:subsampling} shows that evaluating with our subsampling yields similar results to using all ground truth pixels with the same or almost the same ranking -- independent of the error measure.

\section{Conclusion}

With Spring, we present a large, high-resolution, high-detail, computer-generated dataset and benchmark for dense matching.
Spring addresses the increasing performance of recent methods in terms of details, by allowing an adequate assessment in high-detail regions during evaluation.
To this end, it provides 6000 photo-realistic HD stereo frame pairs with 23812 and 12000 super-resolved UHD ground truth frames for motion and stereo, respectively.
Unlike several other benchmarks for matching tasks, Spring not only covers optical flow or disparity estimation:
It is the first benchmark in the deep learning era that also evaluates image-based scene flow, which is essential to enable further progress in this field.
%where so far only the comparably small KITTI benchmark was existing. 
Initial results of 15 non-finetuned baselines show that Spring is a challenging benchmark for recent methods -- particularly with respect to high-detail, non-rigid, unmatched and sky regions.

\paragraphnew{Acknowledgements.}
The Spring movie assets by Blen\-der\linebreak Foundation
%(\url{https://cloud.blender.org/spring}) 
%(\url{cloud.blender.org/spring}) 
$\vert$ \href{https://cloud.blender.org/spring}{cloud.blender.org/spring}
are licensed under CC BY 4.0.
%The authors 
We thank Andy Goralczyk, Francesco Siddi and the %entire 
Blender Studio team.
%L.M.\ 
Lukas Mehl
acknowledges funding %
%Funded 
by the Deutsche Forschungsgemeinschaft (DFG, German Research Foundation) -- Project-ID 251654672 -- TRR 161 (B04).
% J.S.\ 
Jenny Schmalfuss
and 
%A.J.\ 
Azin Jahedi acknowledge support from the International Max Planck Research School for Intelligent Systems
(IMPRS-IS).

%%%%%%%%% REFERENCES
{\small
\bibliographystyle{ieee_fullname}
\bibliography{references}
}

\end{document}